\documentclass{article}

    \PassOptionsToPackage{numbers, compress}{natbib}

\usepackage[preprint]{neurips_2024}

\usepackage{url}
\usepackage{hyperref}

\usepackage{graphicx}
\usepackage{amsfonts}
\usepackage{amsmath}
\usepackage{siunitx}
\usepackage{amssymb}
\usepackage{pifont}
\usepackage{nicefrac}
\usepackage{multirow}
\usepackage{subcaption}
\usepackage[font=small]{caption}
\usepackage{soul}
\usepackage{booktabs}
\usepackage{xcolor}
\usepackage{comment}
\usepackage{sidecap}
\usepackage{xspace}
\usepackage[most]{tcolorbox}
\usepackage{enumitem}
\usepackage{ulem}
\usepackage{makecell}
\usepackage[textsize=tiny,textwidth=24mm]{todonotes}
\usepackage{longtable}
\usepackage{algorithm}
\usepackage{algpseudocode}
\usepackage{wrapfig}

\newsavebox{\tablebox}

\usepackage[utf8]{inputenc}
\usepackage[T1]{fontenc}
\usepackage{CJKutf8}

\definecolor{bblue}{HTML}{4F81BD}
\definecolor{rred}{HTML}{c4260b}
\definecolor{ggreen}{HTML}{098c1f}
\definecolor{ppurple}{HTML}{9F4C7C}
\definecolor{oorange}{HTML}{F79646}

\newcommand{\youtube}{{YouTube}\xspace}
\newcommand{\madlad}{{MADLAD}\xspace}
\newcommand{\fleurs}{{FLEURS}\xspace}

\usepackage{tablefootnote}

\algnewcommand\algorithmicinput{\textbf{Input:}}
\algnewcommand\Input{\item[\algorithmicinput]}
\algnewcommand\algorithmicoutput{\textbf{Output:}}
\algnewcommand\Output{\item[\algorithmicoutput]}

\algnewcommand\algorithmicempty{~}
\algnewcommand\Empty{\item[\algorithmicempty]}

\newcolumntype{H}{>{\setbox0=\hbox\bgroup}c<{\egroup}@{}}

\title{Scaling Sign Language Translation}

\author{%
  Biao Zhang$^1$ \quad Garrett Tanzer$^2$ \quad Orhan Firat$^1$ \\
  $^1$ Google DeepMind \quad $^2$ Google \\
  \texttt{\{biaojiaxing,gtanzer,orhanf\}@google.com} \\
}

\begin{document}

\maketitle

\begin{abstract}

Sign language translation (SLT) addresses the problem of translating information from a sign language in video to a spoken language in text.
Existing studies, while showing progress, are often limited to narrow domains and/or few sign languages and struggle with open-domain tasks.
In this paper, we push forward the frontier of SLT by scaling pretraining data, model size, and number of translation directions.
We perform large-scale SLT pretraining on different data including 1) noisy multilingual \youtube SLT data,
2) parallel text corpora, and 3) SLT data augmented by translating video captions to other languages with off-the-shelf machine translation models.
We unify different pretraining tasks with task-specific prompts under the encoder-decoder architecture, and initialize the SLT model with pretrained (m/By)T5 models across model sizes.
SLT pretraining results on How2Sign and \fleurs-ASL\#0 (ASL to 42 spoken languages) demonstrate the significance of data/model scaling and cross-lingual cross-modal transfer, as well as the feasibility of zero-shot SLT.
We finetune the pretrained SLT models on 5 downstream \textit{open-domain} SLT benchmarks covering 5 sign languages. Experiments show substantial quality improvements over the vanilla baselines, surpassing the previous state-of-the-art (SOTA) by wide margins.

\end{abstract}

\begin{figure*}[h]
\centering
\includegraphics[width=0.55\textwidth]{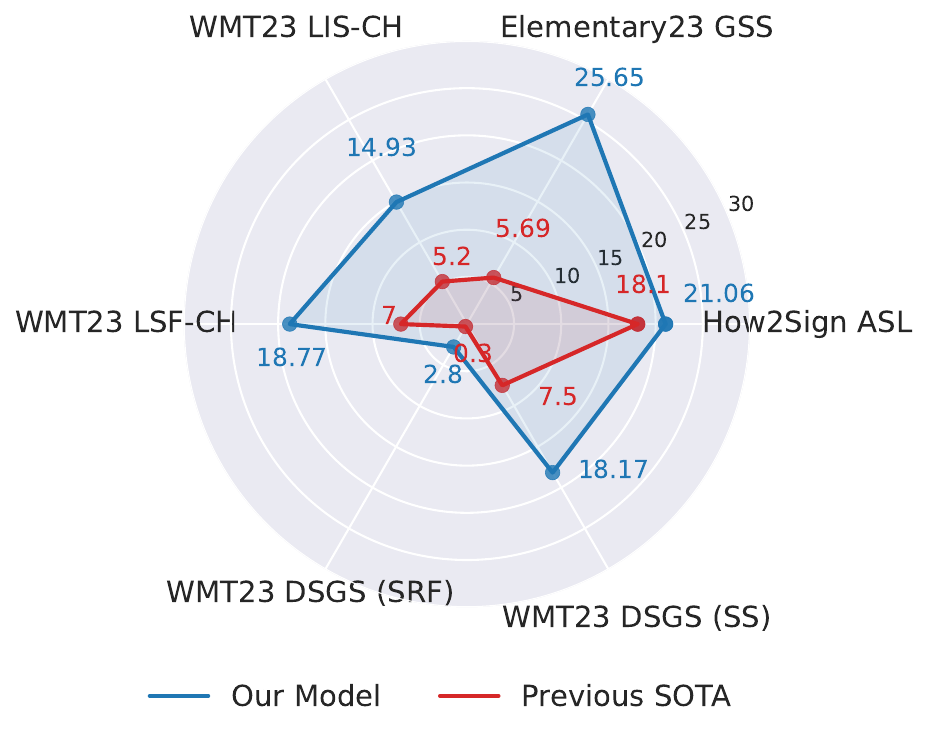}
\caption{\label{fig:sota} BLEU scores on different benchmarks: our model sets new SOTA results across benchmarks and sign languages. Note we didn't show BLEURT because not all previous studies report BLEURT.}
\end{figure*}

\section{Introduction}\label{sec:introduction}

Scalable neural networks trained on large amount of unlabeled and/or weakly-labeled data from multiple modalities and multiple tasks have resulted in performance significantly exceeding that of single-task models trained on particular domains~\citep{achiam2023gpt,dehghani2023scaling,pratap2023scaling,team2023gemini}.
Sign language translation (SLT), as a video-to-text translation task\footnote{While spoken language can be conveyed through either text or speech, this study focuses on text.}, features significant cross-modality challenges in video understanding and text generation.
While extra forms of supervision such as glosses have been helpful in bridging the modality gap~\citep{camgoz2018neural}, they are nonstandardized/incomplete systems available only for small datasets~\citep{desai2024systemic}.
Researchers have instead turned to more scalable approaches such as adapting pretrained vision and text models~\citep{chen2022simple,rust2024towards,wong2024signgpt} and jointly modeling with machine translation~\citep[MT,][]{zhang2023sltunet}.
Despite encouraging progress, these studies were performed at small scale with success on narrowed domains and on few sign languages.
In open-domain SLT settings, unfortunately, they have shown limited effectiveness~\citep{muller-etal-2023-findings}.

\begin{figure*}[t]
\centering
\setlength{\tabcolsep}{4pt}
\small
\subcaptionbox{\label{fig:slt_illustration} Encoder-decoder based SLT model and different SLT pretraining tasks. 
We use red, green, and blue colors to indicate the input prompt, sign frames, and target output, respectively.
``\texttt{sign lang}'': sign language name; ``\texttt{src lang/tgt lang}'': source/target spoken language name; ``\texttt{<$*$>}'': task-specific control tokens; ``\texttt{source/target text}'': source/target text for MT; ``\texttt{clip frames (clip text)}'': concatenation of sign frames (caption texts) corresponding to a video clip; ``\texttt{translated clip text}'': augmented data by off-the-shelf MT models; ``\texttt{clip text with timestamps}'': concatenation of caption texts and their start and end timestamps.}[\textwidth]{
    {\includegraphics[width=0.85\textwidth]{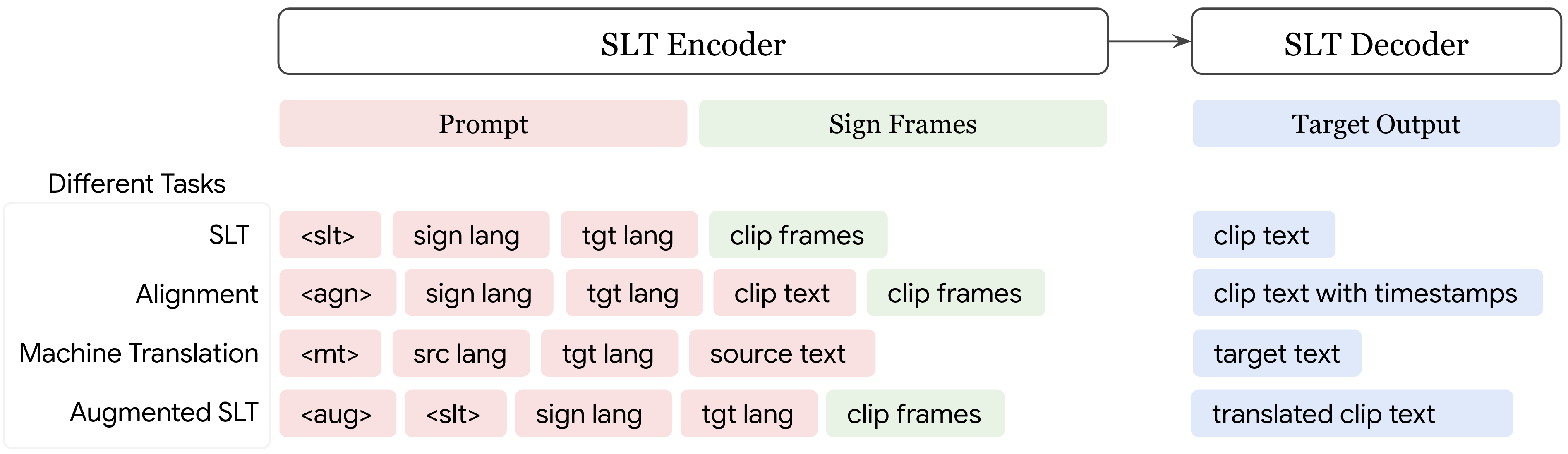}}
}

\vspace{\baselineskip}
\subcaptionbox{\label{fig:clip_structure} Clip overview. Top: a sequence of skeletons for a sign language video where the used keypoints are annotated in red; Bottom: We pretrain SLT models on randomly sampled clips of $N$ seconds from the video. Each segment in the plot represents a caption, and $[\text{Cap}_i, \ldots, \text{Cap}_j]$ (i.e., green segments) denotes captions fully covered by the clip. ``$S_i/E_i$'': the start/end time stamp for caption $\text{Cap}_i$.}[\textwidth]{
\includegraphics[width=0.85\textwidth]{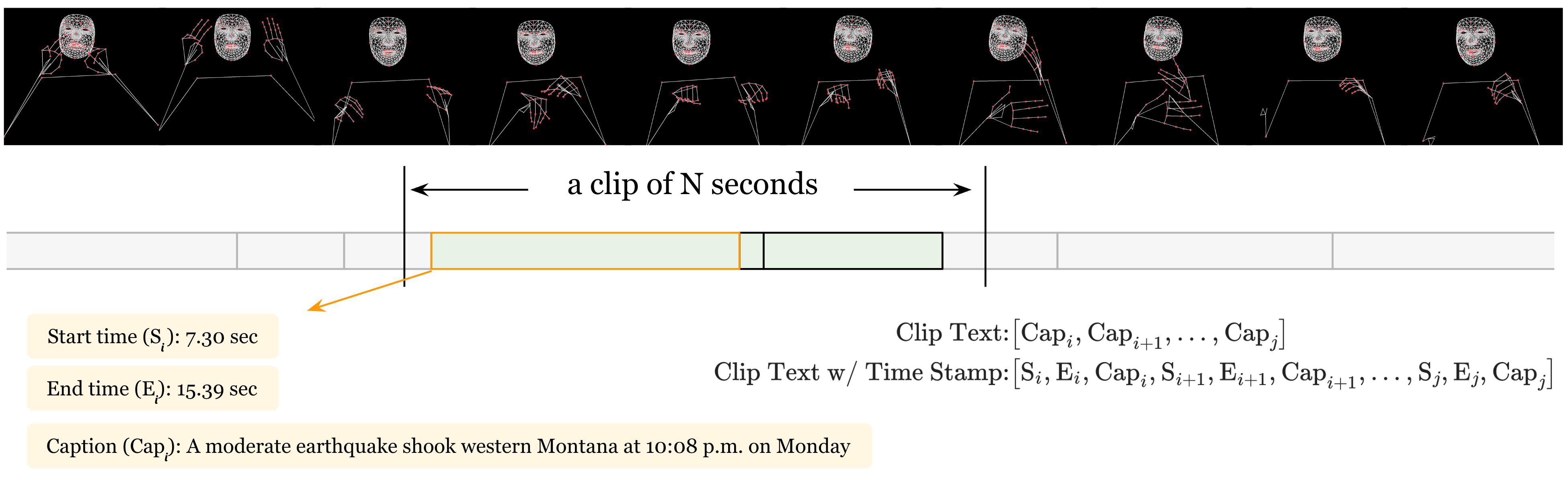}
}

\caption{\label{fig:overview} Illustration of model architecture and pretraining task for SLT. We perform large-scale pretraining and adopt multi-task learning at clip level (multiple captions) to better leverage the supervised knowledge.}
\end{figure*}

In this paper, we aim to improve \textit{open-domain} SLT for multiple sign languages by means of large-scale SLT pretraining with more data, larger models and more languages.
Inspired by the finding that jointly training SLT models with MT data enables positive knowledge transfer to SLT~\citep{zhang2023sltunet}, we explore the following pretraining tasks and data: \textit{web-crawled multilingual SLT}, \textit{multilingual MT}, and \textit{augmented SLT}.
Although high-quality SLT training data are scarce, weakly-labeled SLT data covering diverse topics and signers are readily available from platforms like \youtube.
Prior studies have demonstrated the feasibility of collecting massive \youtube SLT data and its effectiveness on improving SLT~\citep{uthus2024youtube,youtubesl25,fsInSlt}, and we follow this effort in a multilingual setup.
Different from SLT, text-based MT datasets are massive and resource-rich across hundreds of spoken languages~\citep{bapna2022building,nllb2022}.
We explore a subset of \madlad-400~\citep{kudugunta2023madlad} including up to 41 spoken languages for the pretraining.
In addition, we construct synthetic multiway SLT data by translating video captions with an off-the-shelf MT model, which allows us to strengthen direct SLT across more translation directions.
We investigate different ways of mixing these data to exploit weakly supervised SLT knowledge as well as cross-task cross-modality transfer at scale.

As in Figure \ref{fig:overview}, we extend the unified encoder-decoder SLT framework from~\citep{uthus2024youtube,fleursasl,youtubesl25,fsInSlt} with extra tasks and modalities similar to~\citep{zhang2023sltunet}, across different pretrained model families (T5, mT5 and ByT5) and different model sizes.
We distinguish different tasks by carefully designed input prompts that contain task-specific control tokens.
This affords us high flexibility in choosing what tasks and languages to incorporate into the pretraining, easing ablations and the scaling.
We then finetune the pretrained SLT models on downstream SLT benchmarks to refine the learned SLT knowledge.

We evaluate the effect of scaling on 6 open-domain SLT benchmarks across 5 sign languages. \fleurs-ASL\#0~\citep{fleursasl}, built on FLORES-200~\citep{nllb2022}, gives us a testbed to analyze multiway American Sign Language (ASL)-to-X SLT (we examine English and 41 other target languages), while the other benchmarks are for a single language pair.
While pretraining results show the acquired general SLT capability, we also report finetuning results following~\citep{uthus2024youtube}.
Our main findings are below:
\begin{itemize}
    \item Adding more pretraining data, either machine translation or sign language translation data, is a promising way to improve SLT, yielding quality gains of varying degrees.
    \item Zero-shot ASL-to-X translation for language pairs not seen during pretraining is achievable by jointly training on ASL-to-En SLT data and En-to-X MT data. 
    \item Augmenting SLT data by translating target captions to other languages with off-the-shelf MT models substantially improves the translation.
    \item Using larger models is not always helpful: ByT5 Base (582M) often outperforms XL (3.74B), but model scaling does benefit SLT when modeling capacity becomes a bottleneck (e.g., when more languages and data are used).
    \item Learned metrics (e.g., BLEURT) show higher correlation between pretrained and finetuned SLT scores than classical metrics (e.g., BLEU or ChrF).
\end{itemize}
Putting everything together, our model achieves new state-of-the-art results across the benchmarks as shown in Figure \ref{fig:sota}, demonstrating the significance of scaling SLT.

\section{Sign Language Translation}\label{sec:slt}

\subsection{Modeling}

We build on a line of work using T5 model families for SLT~\citep{uthus2024youtube,fleursasl,youtubesl25,fsInSlt}, which build upon earlier SLT work~\citep{camgoz2018neural,zhou2021spatial,zhang2023sltunet} using the encoder-decoder architecture~\citep{NIPS2014_a14ac55a,vaswani2017attention}. 
Figure \ref{fig:slt_illustration} shows the overall structure.
The encoder takes as input the concatenation of a prompt instructing the task and a sequence of sign language video frames; the decoder predicts the text output in a target spoken language one token at a time.
We adopt the family of pretrained (m/By)T5 models~\citep{raffel2020t5,xue-etal-2022-byt5} as the backbone and adapt them to SLT via large-scale SLT pretraining followed by downstream finetuning, i.e. \textbf{(m/By)T5 initialization} $\rightarrow$ \textbf{SLT pretraining} $\rightarrow$ \textbf{SLT finetuning}.

We rely on web-crawled \youtube SLT data for SLT pretraining, which provide high coverage on domains and signers albeit at lower quality.
Although recent debates value data quality over data quantity in pretraining~\citep{gunasekar2023textbooks,lee-etal-2022-deduplicating,nguyen2022quality}, we argue that they were established on the availability of massive high-quality training data, which doesn't hold for SLT yet.
We expect that the pretraining could capture the (weakly) supervised SLT knowledge from the crawled data as in previous studies~\citep{uthus2024youtube}.

As shown in Figure \ref{fig:clip_structure}, we adopt the clip-level training following~\citep{fleursasl} that randomly samples a clip of $N$ seconds from the sign video and then predicts various types of in-clip information (such as caption texts and their start and end timestamps) based on the frames of the entire clip.
Detailed tasks are listed in Figure \ref{fig:slt_illustration}, which are all formulated as sequence-to-sequence tasks.
They are distinguished by prompts with different control tokens and are trained with the standard maximum likelihood objective.
For the \textbf{baseline}, we consider the following two tasks: \textit{SLT} and \textit{alignment}, and train it by mixing these two tasks with a pre-specified mix ratio.
\begin{description}
    \item[{SLT}]
    This is the core task that directly models the translation from clip frames to the clip text in a target language.
    It is indispensable for the model to acquire the translation capability.
    \item[{Alignment}]
    It is an auxiliary task for SLT, learning to align the input clip with its captions.
    We train the model to infer the start and end time stamp for each in-clip caption.
    Apart from regularization, this task could improve the model's understanding of sign language~\citep{fleursasl}.
\end{description}

\subsection{Scaling Model Size, Number of Languages and Pretraining Data Size}

\textbf{Model Scaling}
Scaling model size increases modeling capacity, which has been widely proven effective in improving the task performance~\citep{achiam2023gpt,team2023gemini,team2024gemini}.
We study whether and how increasing model size affects the SLT performance and compare (By/m)T5 models for SLT at different scales.

\textbf{Language Scaling}
While most SLT works focus on a few sign and spoken languages, we expand our study to massive languages, covering up to 80 sign/spoken languages during pretraining, and 5 sign language and 42 spoken languages at evaluation.
We are interested in whether a single SLT model could support multiple sign/spoken languages with non-trivial performance, and whether knowledge transfer could improve SLT on low-resource languages~\citep{zhang-etal-2020-improving,youtubesl25}.

\textbf{Data Scaling}
Data scarcity is the main bottleneck hindering the development of SLT. To address this issue, we investigate the following three types of data for the pretraining:
\begin{description}
    \item[{SLT}]
    We crawl multilingual \youtube SLT data following the recipe~\citep{youtubesl25} except that we didn't perform human annotation and filtering.
    This allows us to significantly scale up the SLT data by 3$\sim$6 times, reaching $\sim$6,600 hours in total, albeit at much lower quality.
    \item[{Machine Translation}]
    Unlike SLT, MT is a text-to-text translation task with rich parallel resources, particularly for high-resource languages~\citep{arivazhagan2019massively}.
    We explore adding multilingual MT data into the pretraining and mark this task with control token ``\texttt{<mt>}''~\citep{zhang2023sltunet}.
    \item[{Augmented SLT}]
    SLT data are often one-to-one translation data, where each sign language only has translation in one spoken language.
    This makes the translation of a sign language to other spoken languages difficult.
    We thus augment SLT data to one-to-many by translating the target text to other spoken languages via off-the-shelf MT models.
    As in Figure \ref{fig:slt_illustration}, we use ``\texttt{<aug>}'' to separate genuine SLT data from the augmented one~\citep{caswell-etal-2019-tagged}.
\end{description}

\section{Setup}\label{sec:setup}

\textbf{MT Pretraining Data}
We use the parallel sentence-level portion of \madlad-400~\citep{kudugunta2023madlad} as the MT pretraining data.
We extract a subset of \madlad-400 for experiments, including 41 languages (apart from English (En)) covering diverse language families and scripts, and explore the impact of En$\rightarrow$Xx and Xx$\rightarrow$En MT data on SLT in experiments.
We create two settings for the pretraining:
\begin{itemize}
    \item \textbf{MT-Small}: A high/medium-resource subset including 11 languages \texttt{es, de, fr, it, cs, pl, ru, zh, ar, ja, hi}.
    \item \textbf{MT-Large}: This set includes all 41 languages. Apart from MT-Small, it has \texttt{nl, pt, sv, hu, da, fi, el, sk, no, bg, lt, lv, sl, et, ko, hr, is, sr, tr, vi, id, he, th, ms, uk, ca, ta, fil, ne, cy}.
\end{itemize}
Table \ref{tab:lang_stats} shows the statistics for each language.
Unless otherwise specified, we balance the MT data distribution over languages during training by temperature sampling with a rate of 5~\citep{arivazhagan2019massively}.

\textbf{SLT Pretraining Data}
We experiment with noisy captioned sign language videos from \youtube. This is the full set of videos pre-manual filtering in~\citep{youtubesl25}.
Estimated statistics for each sign language are summarized in Table \ref{tab:lang_stats}.
We also have two settings for this data:
\begin{itemize}
    \item \textbf{YT-ASL}: $\sim$2,800 hours of noisy captioned ASL videos; a superset of \youtube-ASL~\citep{uthus2024youtube} (modulo video churn) and the same dataset used by~\citep{fsInSlt}.
    \item \textbf{YT-Full}: $\sim$6,600 hours of noisy captioned multilingual sign language videos; a superset of~\citep{youtubesl25}.
\end{itemize}
During training, we mix the SLT data for all languages in proportion to their duration.
We further augment these data with other spoken languages via \madlad-MT-3B~\citep{kudugunta2023madlad}.
For ASL SLT data, we translate the English captions to 41 spoken languages listed in MT-Large, which makes YT-ASL 43-way multilingual SLT, namely \textbf{Aug-YT-ASL}; 
for other SLT data, we translate the target text into English, resulting in 3-way multilingual SLT.\footnote{Note translations were performed per caption, which may lack coherence when compiled into a document.}
We refer to the augmented SLT data for all sign languages as \textbf{Aug-YT-Full}. 
Similar to MT-Small and MT-Large, we reorganize the augmented data to \textbf{Aug-YT-ASL-Small/Aug-YT-Full-Small} and \textbf{Aug-YT-ASL-Large/Aug-YT-Full-Large}.

\textbf{SLT Pretraining Mixture}
We ablate across several SLT pretraining mixtures.
\begin{itemize}
    \item \textbf{Baseline}: Caption alignment and SLT tasks. We use the task weights from~\citep{fleursasl}, including $4\%$ for alignment.
    \item \textbf{Baseline + MT}: We mix MT data into Baseline with a sampling probability of $p_{MT}$.
    \item \textbf{Baseline + Augmented SLT}: We replace the Baseline SLT data with the augmented SLT data and uniformly sample the target language for each example at each step.
    \item \textbf{Baseline + MT + Augmented SLT}: \textit{Baseline + MT} but with augmented target languages, as above.
\end{itemize}

\begin{table*}[t]
\centering
\setlength{\tabcolsep}{4pt}
\small
\begin{tabular}{lccrrr}
\toprule
Task & Sign Lang & Target Lang & \#Train & \#Dev & \#Test  \\
\cmidrule(lr){1-6}
How2Sign & ASL & En & 183,097 & 10,277 & 13,890 \\
Elementary23 & GSS & El & 35,970 & 512 & 512 \\
\multirow{3}{*}{WMT23}
& LIS-CH & It & 1,901 & 100 & 250 \\
& LSF-CH & Fr & 5,560 & 100 & 250 \\
& DSGS & De & 310,840 & \begin{tabular}{@{}r@{}}420 \\ (WMT22)\end{tabular} & \begin{tabular}{@{}r@{}}250/246 \\ (SS/SRF split)\end{tabular} \\
\cmidrule(lr){1-6}
\fleurs-ASL\#0 & ASL & 200 Flores Langs & - & - & 353 \\ 
\bottomrule
\end{tabular}
\caption{\label{tab:downstream_tasks} Summary of downstream SLT benchmarks. ``\#Train/\#Dev/\#Test'': the number of examples in the train, dev and test split. Note the sign language video and the target text in these benchmarks are often pre-segmented and aligned at sentence level. ``DGS/ASL/GSS'': German/American/Greek Sign Language; ``En/De/Fr/It'': English/German/French/Italian; ``LIS-CH'': Italian Sign Language of Switzerland; ``LSF-CH'': French Sign Language of Switzerland; ``DSGS'': Swiss German Sign Language.}
\end{table*}

\textbf{Downstream Benchmarks, Evaluation and Model Setting}
We thoroughly evaluate the translation performance on a range of \textit{open-domain} SLT benchmarks, including 
How2Sign~\citep{Duarte_CVPR2021}, Elementary23~\citep{voskou2023new}\footnote{While not as restricted as specific domains like ``weather forecasts'', the scope of topics in Elementary23 remains somewhat focused.}, WMT23~\citep{muller-etal-2023-findings} and \fleurs-ASL\#0 (signer id \#0)~\citep{fleursasl}.
Detailed information for each benchmark is given in Table \ref{tab:downstream_tasks}.
Overall, the evaluation covers 5 source sign languages and 42 target spoken languages.\footnote{We acknowledge that there are other SLT benchmarks available in academia. We didn't include them in our experiments due to their licensing restrictions and/or domain limitations.}


We report translation results for \textbf{Pretraining} and \textbf{Finetuning}.
During inference, we use beam search with a beam size of 5.
We evaluate translation with detokenized BLEU~\citep{bleu2002} and ChrF~\citep{popovic-2015-chrf}, as well as neural metric, BLEURT~\citep{pu2021learning}.
We use BLEURT as the main metric~\citep{freitag-etal-2022-results}.
We initialize our SLT model with three T5 model families: T5~\citep{raffel2020t5}, mT5~\citep{xue2020mt5} and ByT5~\citep{xue-etal-2022-byt5}, at three different sizes: Base, Large and XL. We optimize models with Adafactor~\citep{shazeer2018adafactor}, and set the maximum text input, landmark input, and text output length to 512.
More setup details are given in Appendix \ref{app:model_setup}.

\section{Experiments}\label{sec:experiments}

\subsection{SLT Pretraining Results}

\begin{table*}[t]
\centering
\setlength{\tabcolsep}{4pt}
\small
\begin{tabular}{llrrrrrr}
\toprule
\multirow{2}{*}{SLT Data} & \multirow{2}{*}{Model}
& \multicolumn{3}{c}{How2Sign} & \multicolumn{3}{c}{\fleurs-ASL\#0 (En)} \\
\cmidrule(lr){3-5} \cmidrule(lr){6-8}
& & Base & Large & XL & Base & Large & XL \\
\cmidrule(lr){1-8}
\multirow{3}{*}{YT-ASL}
& T5   & \textbf{29.54} & 27.95 & 22.96 & \textbf{32.8} & 4.18 & 32.41 \\
& mT5  & \textbf{34.94} & 8.46 & 23.7 & 35.59 & \textbf{43.53} & 23.09 \\
& ByT5 & \textbf{30.36} & 23.51 & 29.2 & \textbf{44.84} & 28.47 & 41.65 \\
\cmidrule(lr){1-8}
\multirow{3}{*}{YT-Full}
& T5   & \textbf{31.64} & 25.45 & 8.57 & \textbf{42.86} & 37.55 & 30.02 \\
& mT5  & \textbf{31.46} & 19.37 & 24.46 & \textbf{38.03} & 24.56 & 33.16 \\
& ByT5 & \textbf{37.13} & 22.61 & 29.59 & 52.48 & 43.01 & \textbf{52.71} \\
\bottomrule
\end{tabular}
\caption{\label{tab:t5_size_cmp} Pretraining performance (BLEURT $\uparrow$) for different sized (By/m)T5 models when pretrained on YT-ASL and YT-Full. Results are reported on the test set of How2Sign and \fleurs-ASL\#0 ($\rightarrow$En, i.e. English as the target). Best results for each model family are highlighted in bold.}
\end{table*}

\textbf{Model scaling doesn't improve SLT consistently: Base often outperforms Large/XL.}
Table \ref{tab:t5_size_cmp} also shows that scaling up model size rarely results in consistent quality improvements. 
Different from findings on text-only tasks~\citep{raffel2020t5,xue-etal-2022-byt5}, Base surpasses Large and XL in most cases, where Large often converges the slowest and performs the worst. 
Model scaling alone doesn't significantly reduce the video-text modality gap, although better optimization and checkpoint selection could help.
XL performs relatively comparable to Base.
When MT data is mixed in and modeling capacity becomes the bottleneck, the value of model scaling by XL emerges as shown in Figure \ref{fig:mt_transfer_distil} and Table \ref{tab:downstream_slt_results}.

\textbf{Backbone affects SLT substantially; ByT5 generally performs the best.}
While several previous studies selected T5~\citep{uthus2024youtube,muller-etal-2023-findings} or mT5~\citep{youtubesl25} as the SLT backbone, we observe in Table \ref{tab:t5_size_cmp} that ByT5-based SLT outperforms its T5 counterpart in most settings, confirming the results of~\citep{fsInSlt} at scale. Given that larger models do not consistently perform better, it seems less likely that ByT5's superiority comes from its encoder-heavy parameter allocation, and more likely that it is due to its spelling capabilities and reduced input length gap between byte text sequences and video frame sequences. Unless otherwise stated, we use ByT5 Base for the following experiments.

\textbf{Scaling SLT data generally improves quality significantly.}
Adding more SLT data, i.e. from YT-ASL to YT-Full, largely improves the translation quality in most settings.
For ByT5-based SLT particularly, the gain reaches $\sim$7 BLEURT on How2Sign and $\sim$11 BLEURT on \fleurs-ASL\#0 (En) for Base and XL, respectively.
We conjecture that adding more (multilingual) SLT data helps reduce the modality gap (especially with skeletons, which lack pretrained representations) and enable cross-lingual knowledge transfer~\citep{arivazhagan2019massively,zhang-etal-2020-improving, 9878501, youtubesl25}.

\begin{figure*}[t]
\centering
\small

\subcaptionbox{\label{fig:mt_transfer_2en_floresen} BLEURT scores for \fleurs-ASL\#0 ($\rightarrow$En).}[0.495\textwidth]{
\includegraphics[width=0.47\textwidth]{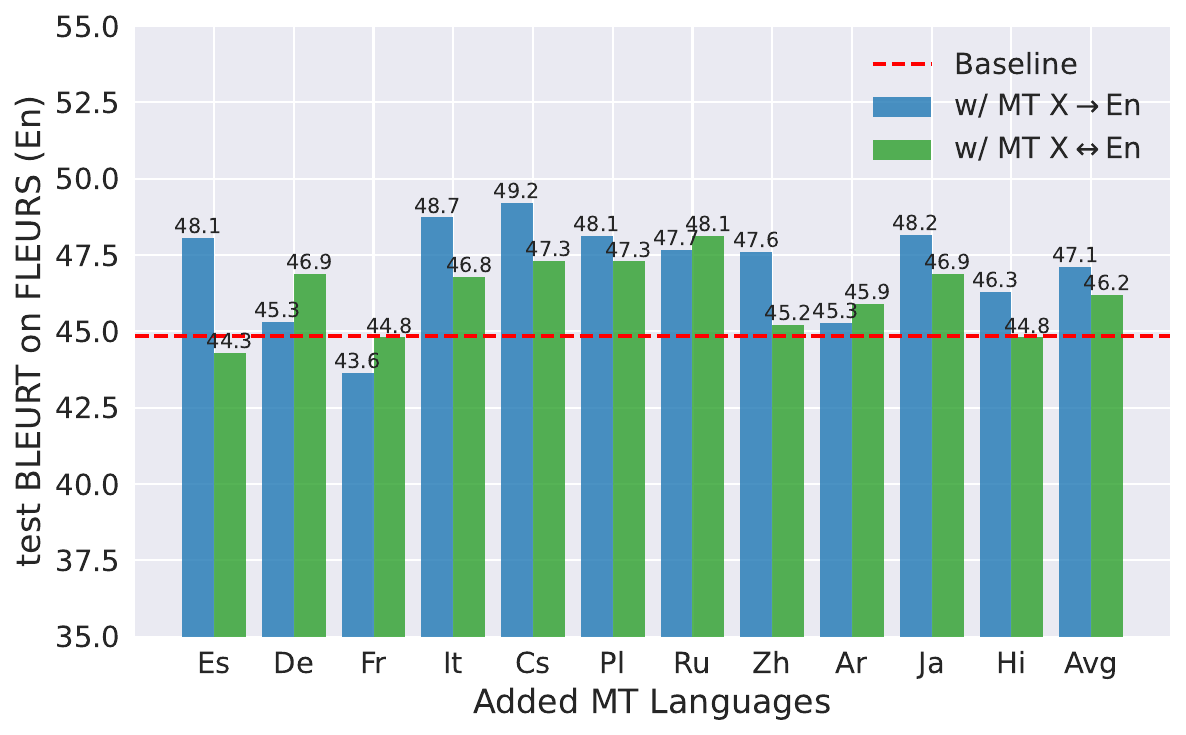}
}
\subcaptionbox{\label{fig:mt_transfer_2xx_floresxx} Zero-shot BLEURT for \fleurs-ASL\#0 ($\rightarrow$X).}[0.495\textwidth]{
\includegraphics[width=0.47\textwidth]{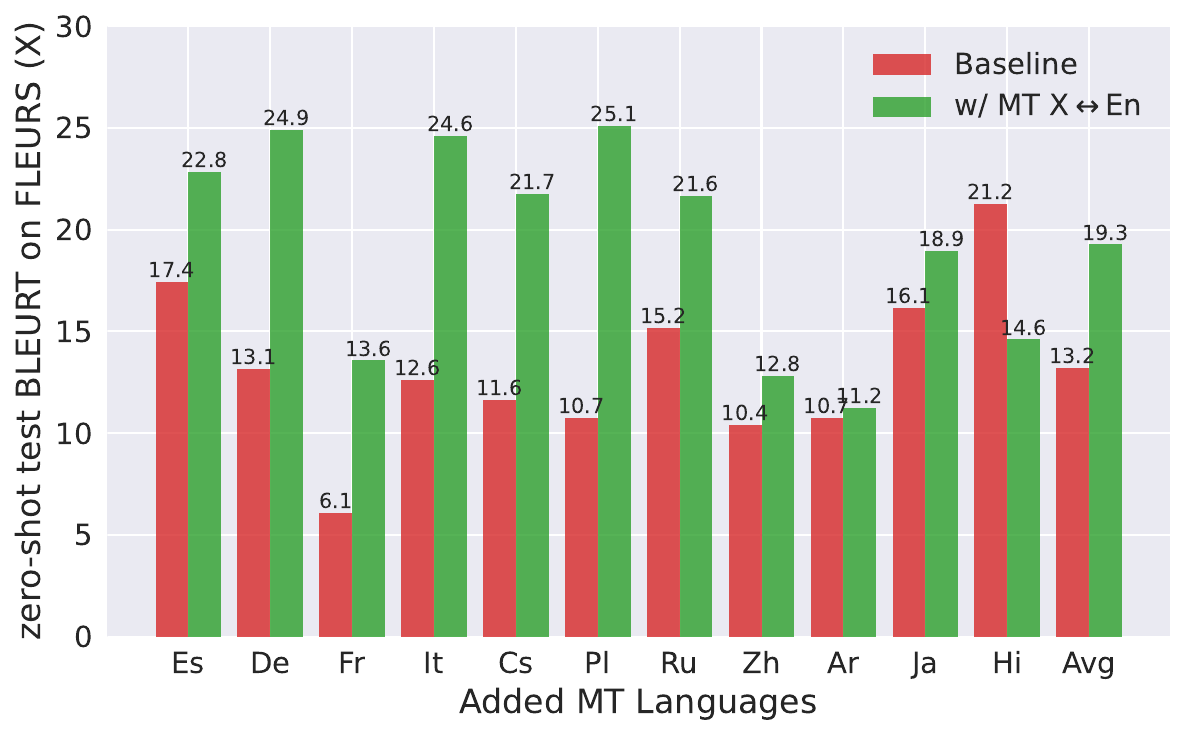}
}

\caption{\label{fig:mt_transfer_lang_impact} Pretraining performance for \textit{Baseline + MT} when varying MT languages. We show BLEURT$\uparrow$ results on \fleurs-ASL\#0, and set $p_{mt}=0.5$. Note MT languages are added separately instead of jointly. Results are for ByT5 Base. 
``X$\rightarrow$En'': MT data for translation into English; ``X$\leftrightarrow$En'': MT data for both translation directions; ``Avg'': average performance over languages. 
MT languages are arranged in descending order from left to right based on their training data quantity.}
\end{figure*}

\textbf{Mixing MT and SLT data yields positive knowledge transfer to SLT.}
We next explore whether and how the addition of MT data benefits SLT, starting with YT-ASL and bilingual MT data with $p_{mt}=0.5$.
Figures \ref{fig:mt_transfer_2en_floresen} and \ref{fig:mt_transfer_2en_how2sign} show that adding bilingual translation data improves SLT performance generally, confirming the findings of SLTUNet~\citep{zhang2023sltunet}---that jointly training with MT enables positive knowledge transfer---at scale.
The quality gains vary greatly across languages, which show little correlation with language family or training data scale.
For example, adding a large amount of Fr$\rightarrow$En data ($\sim$243M sentence pairs) helps little (or even hurts) on \fleurs-ASL\#0 (En), while adding a small amount of Ja$\rightarrow$En data ($\sim$5M sentence pairs) gives a gain of at least $3$ BLEURT on How2Sign and \fleurs-ASL\#0 (En).

\textbf{Translation direction of MT data affects transfer to SLT.}
There are three ways to leverage MT data for SLT: 1) X$\rightarrow$En, 2) En$\rightarrow$X, and 3) both.
We compare 1) and 3) in Figures \ref{fig:mt_transfer_2en_floresen} and \ref{fig:mt_transfer_2en_how2sign} for ASL-to-En SLT.
The translation direction of MT data influences SLT performance greatly and varies across languages.
On average, X$\rightarrow$En benefits ASL-to-En SLT more than X$\leftrightarrow$En: +0.08 and +0.9 BLEURT on How2Sign and \fleurs-ASL\#0 (EN), respectively.
We speculate that including translation into X uses model capacity, which, while enabling zero-shot ASL-to-X SLT as discussed below, results in slightly worse ASL-to-En performance.
This suggests that MT data with the same target language as SLT is most effective for transfer.
Table \ref{tab:mt_transfer_m4_size} shows further support where En$\rightarrow$X surpasses X$\rightarrow$En on multilingual SLT.

\begin{wraptable}{r}{0.40\textwidth}
\centering
\small

    \resizebox{0.40\textwidth}{!}{
    \begin{tabular}{llrr}
    \toprule
    \multirow{2}{*}{SLT Data} & \multirow{2}{*}{Dir} & \multicolumn{2}{c}{BLEURT} \\
    \cmidrule(lr){3-4}
    & & Small & Large \\
    \cmidrule(lr){1-4}
    \multicolumn{2}{c}{Baseline + YT-ASL} & 15.85 & 17.21 \\
    \multicolumn{2}{c}{Baseline + YT-Full} & 24.36 & 23.16 \\
    \cmidrule(lr){1-4}
    \multicolumn{3}{l}{\textit{Baseline + MT-Small}} \\
    \multirow{3}{*}{YT-ASL}
    & En$\rightarrow$X     & 23.51 & 21.25 \\
    & X$\rightarrow$En     & 17.44 & 19.72 \\
    & En$\leftrightarrow$X & 23.84 & 19.19 \\
    \cmidrule(lr){2-4}
    \multirow{3}{*}{YT-Full}
    & En$\rightarrow$X     & 27.29 & 23.15 \\
    & X$\rightarrow$En     & 22.48 & 22.47 \\
    & En$\leftrightarrow$X & 26.33 & 21.48 \\
    \cmidrule(lr){1-4}
    \multicolumn{3}{l}{\textit{Baseline + MT-Large}} \\
    YT-ASL & En$\leftrightarrow$X & 24.69 & 26.60 \\
    YT-Full & En$\leftrightarrow$X & \textbf{29.52} & \textbf{30.69} \\
    \bottomrule
    \end{tabular}
    }

\caption{\label{tab:mt_transfer_m4_size} Pretraining performance for \textit{Baseline + MT} with $p_{mt}=0.9$ when scaling up languages and data. We show averaged BLEURT$\uparrow$ results on \fleurs-ASL\#0. 
Results are for ByT5 Base. ``Dir'': translation direction of MT data; ``Small/Large'': average results over the target languages included in MT-Small/MT-Large on \fleurs-ASL\#0.}
\end{wraptable}

\textbf{We can achieve zero-shot bilingual ASL-to-X SLT via ASL-to-En SLT + En$\leftrightarrow$X MT, albeit at poor quality.}
If knowledge can be transferred from MT to SLT, one straightforward question is whether we can achieve zero-shot SLT by jointly training with MT.
We do so by training on ASL-to-En SLT + En$\leftrightarrow$X MT data and examining zero-shot ASL-to-X SLT on \fleurs-ASL\#0 (X).
Figure \ref{fig:mt_transfer_2xx_floresxx} shows that this works effectively.
On Pl and It, we observe quality gains over 12 BLEURT; on average, adding MT data improves zero-shot SLT by $\sim$6 BLEURT.
Nevertheless, the overall zero-shot SLT performance is middling, and the gains are unstable across languages, e.g. performance degrades for ASL-to-Hi SLT with joint MT training.
Similar findings were also observed in multilingual MT and speech translation~\citep{zhang-etal-2020-improving,dinh2021zero}.

\begin{figure*}[t]
\centering
\small

\includegraphics[width=0.98\textwidth]{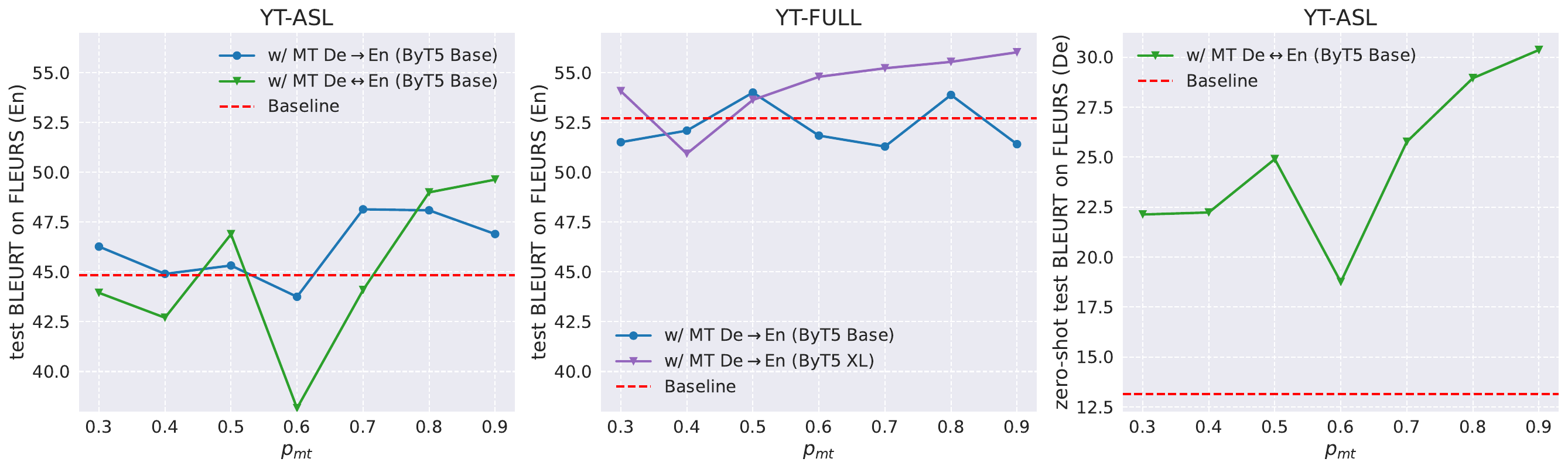}

\caption{\label{fig:mt_transfer_ratio} Pretraining performance for \textit{Baseline + MT} when changing the mixing ratio of MT data $p_{mt}$ on \fleurs-ASL\#0 (En and De) test set. We show BLEURT$\uparrow$ results as we vary $p_{mt}$ from 0.3 to 0.9.
}
\end{figure*}

\textbf{Using a higher sampling ratio for the MT data, i.e. larger $p_{mt}$, often improves SLT.}
We start with $p_{mt}=0.5$, i.e., sampling equal amount of SLT and MT data, in the above experiments following intuition.
However, the proportion of different types of data often has non-negligible influence in multilingual modeling~\citep{arivazhagan2019massively,conneau2019unsupervised}.
We next explore its impact on SLT and use MT En-De for illustration.
Figure \ref{fig:mt_transfer_ratio} and \ref{fig:mt_transfer_ratio_how2sign} shows that $p_{mt}=0.5$ is sub-optimal and sampling more MT data improves SLT in most settings, regardless of using ByT5 Base or XL, YT-ASL or YT-Full, MT De$\rightarrow$En or De$\leftrightarrow$En, and How2Sign or \fleurs-ASL\#0 (En/De).
In addition, increasing the proportion of MT data also improves zero-shot ASL-to-De SLT.
Note another benefit of using more MT data is to accelerate training, as loading SLT data is much slower than loading text-only MT data.
We use $p_{mt}=0.9$ by default in the following experiments.

\textbf{Multilingual MT improves multilingual (zero-shot) SLT.}
The above experiments mainly analyze SLT with bilingual MT.
We next investigate how multilingual MT affects multilingual (zero-shot) SLT, particularly the use of MT-Small and MT-Large.
We report results for ASL-to-\textit{Small} and ASL-to-\textit{Large} SLT on \fleurs-ASL\#0 where Small and Large denote the target languages covered by MT-Small and MT-Large, respectively.
Note all SLT directions are zero-shot except the translation to English.

Table \ref{tab:mt_transfer_m4_size} summarizes the average performance.
Using multilingual X$\rightarrow$En MT data results in unstable ASL-to-X SLT performance, which even hurts SLT on YT-Full.
In contrast, multilingual En$\rightarrow$X and En$\leftrightarrow$X MT data are both very helpful to SLT, where the former often outperforms the latter.
By default, we still use En$\leftrightarrow$X MT data in the following experiments so as to fully leverage the knowledge in MT data during pretraining.

Figures \ref{fig:mt_transfer_m4_small} and \ref{fig:mt_transfer_m4_large} further show the language breakdown results.
Adding multilingual MT significantly improves ASL-to-En SLT when using YT-ASL alone, while the gain almost disappears when using larger-scale SLT data, YT-Full.
Again, we note that the overall zero-shot translation quality is poor -- the best average BLEURT on Small and Large is 29.52 and 30.69, respectively.
Achieving significant ASL-to-X SLT requires techniques beyond naive SLT and MT data mixing.

\begin{wraptable}{r}{0.45\textwidth}
\centering
\small

    \begin{tabular}{lrr}
    \toprule
    \multirow{2}{*}{Setting} & \multicolumn{2}{c}{BLEURT} \\
    \cmidrule(lr){2-3}
    & Small & Large \\
    \cmidrule(lr){1-3}
    Baseline + YT-ASL & 15.85 & 17.21 \\
    + Aug-YT-ASL-Small & 31.14 & 19.74 \\
    \quad + MT-Small & 30.51 & 19.70 \\
    + Aug-YT-ASL\&MT-Large & 25.83 & 33.71 \\
    \quad + ByT5 XL & \textbf{38.53} & \textbf{42.56} \\
    + MT-3B Cascading & 34.82 & 37.82 \\
    \cmidrule(lr){1-3}
    Baseline + YT-Full & 24.36 & 23.16 \\
    + Aug-YT-Full-Small & 38.53 & 29.49 \\
    \quad + MT-Small & 36.84 & 25.67 \\
    + Aug-YT-Full\&MT-Large & 36.01 & 39.85 \\
    \quad + ByT5 XL & \textbf{45.12} & \textbf{48.05} \\
    + MT-3B Cascading & 43.54 & 46.32 \\
    \quad + ByT5 XL & 44.82 & 47.52 \\ 
    \bottomrule
    \end{tabular}

\caption{\label{tab:mt_transfer_distill} Pretraining performance (averaged BLEURT$\uparrow$) for \textit{Baseline + Augmented SLT + MT} with $p_{mt}=0.9$ on \fleurs-ASL\#0 test set. MT data are multilingual in both directions.
Baseline is for ByT5 Base; ``MT-3B'': \madlad-MT-3B, the model used for SLT augmentation; ``Cascading'': translating \fleurs-ASL\#0 to English and then performing MT to other target languages.}
\end{wraptable}

\textbf{Data augmentation and large-capacity modeling are promising methods for multilingual SLT.}
In MT, a common solution to improve zero-shot quality is to construct pseudo translation data for zero-shot directions~\citep{arivazhagan2019massively,fan2021beyond,zhang-sennrich-2021-edinburghs,freitag2020complete}.
We examine this practice for SLT.
We adopt publicly pretrained MT models to generate data for more target languages for the \youtube SLT data (i.e., Augmented SLT).
Results in Table \ref{tab:mt_transfer_distill} demonstrate the effectiveness of Augmented SLT, which significantly improves the best performance for ByT5 Base-based SLT to 36.01 and 39.85 average BLEURT on Small and Large with a gain of 6.49 (29.52$\rightarrow$36.01) and 9.16 (30.69$\rightarrow$39.85), respectively.
Note there are 42 languages in Large.
ByT5 Base may be insufficient in accommodating translation for such amount of languages.
Increasing the modeling capacity to XL yields another gain of 9.11 (36.01$\rightarrow$45.12) and 8.2 (39.85$\rightarrow$48.05) average BLEURT on Small and Large, respectively.
On YT-ASL, Augmented SLT and ByT5 XL also lead to substantial quality improvements by 13.84 (24.69$\rightarrow$38.53)/15.96 (26.60$\rightarrow$42.56) average BLEURT on Small/Large.
The final performance even surpasses the cascading baseline, i.e. ASL-to-En SLT chained with En-to-X MT, under both YT-ASL and YT-Full.
Figures \ref{fig:mt_transfer_m4_small_distil} and \ref{fig:mt_transfer_m4_large_distil} also show the quality improvements across languages resulted from data augmentation and ByT5 XL.

\begin{table*}[t]
\centering
\setlength{\tabcolsep}{4pt}
\small

\resizebox{0.85\textwidth}{!}{\begin{tabular}{llrrrrrrr}
\toprule
\multirow{2}{*}{ID} & \multirow{2}{*}{Model} & \multirow{2}{*}{H2S} & \multirow{2}{*}{E23} & \multicolumn{4}{c}{WMT23} & \multirow{2}{*}{Avg} \\
\cmidrule(lr){5-8}
& & & & LIS-CH & LSF-CH & SRF & SS & \\
\cmidrule(lr){1-9}
0 & Prevous SOTA & 50.80 & - & 25.20 & 18.80 & 24.60 & 37.70 & - \\
\cmidrule(lr){1-9}
1 & ByT5 Base & 34.00 & 22.14 & 22.77 & 7.74 & 15.41 & 26.88 & 21.49 \\
2 & 1 + Baseline + YT-ASL & 51.74 & 37.79 & 24.24 & 15.43 & 21.82 & 35.59 & 31.10 \\
3 & 2 + MT-Small ($p_{mt}=0.9$) & 52.62 & 45.98 & 33.10 & 24.58 & 23.33 & 45.45 & 37.51 \\
4 & 3 + Aug-YT-ASL-Small & 53.36 & 49.34 & 38.61 & 28.70 & 25.87 & 49.61 & 40.91 \\
5 & 4 + Aug-YT-ASL\&MT-Large + ByT5 XL & 54.28 & 54.16 & 38.93 & 27.29 & 28.42 & 51.73 & 42.47 \\
\cmidrule(lr){1-9}
6 & 2 + YT-Full & 53.51 & 49.48 & 42.11 & 31.16 & 21.15 & 44.28 & 40.28 \\
7 & 6 + Aug-YT-ASL\&MT-Small & 53.70 & 53.13 & 45.09 & 37.69 & 30.31 & 52.45 & 45.40 \\
8 & 7 + Aug-YT-ASL\&MT-Large + ByT5 XL & \textbf{55.69} & \textbf{56.94} & 51.94 & \textbf{41.14} & \textbf{33.94} & 57.96 & \textbf{49.60} \\
\cmidrule(lr){1-9}
9 & 8 + Multilingual SLT Tuning & 53.47 & 55.57 & \textbf{54.54} & 39.26 & 29.33 & \textbf{58.08} & 48.38 \\
\bottomrule
\end{tabular}}

\caption{\label{tab:downstream_slt_results} Finetuning performance (BLEURT$\uparrow$) on downstream SLT benchmarks. ``H2S/E23'': How2Sign/Elementary23. ``SRF/SS'': WMT23 DSGS SRF/SS test split. ``Avg'': averaged performance over all benchmarks. MT data are added in both translation directions. Previous SOTA: How2Sign~\citep{fsInSlt}, Elementary23~\citep{voskou2023new} and WMT23 SRF~\citep{muller-etal-2023-findings}, WMT23 LIS-CH, LSF-CH, SS~\citep{youtubesl25}. \textit{All models are finetuned on each SLT benchmark separately except (9).}}
\end{table*}

\subsection{SLT Finetuning Results}


\textbf{Finetuning on downstream benchmarks substantially improves SLT performance.}
Table \ref{tab:downstream_slt_results} shows that finetuning the pretrained SLT models yields substantial quality gains across benchmarks and settings.
This is because the potential of pretrained models is not fully elicited by direct evaluation due to video recording, domain and (clip-based) pretraining vs. (segment-based) inference mismatches, and finetuning largely mitigates these gaps.
For example, pretraining with external augmented SLT and MT data results in even worse pretraining performance ((6)$\rightarrow$(7)) in Table \ref{tab:downstream_slt_pretrain_results}.
After finetuning, nevertheless, model (7) significantly surpasses model (6) by 5.12 BLEURT on average.

Adding multilingual SLT data (YT-Full) into the pretraining greatly improves the performance from 14.26 (model (5)) to 32.48 BLEURT (model (8)) in Table \ref{tab:downstream_slt_pretrain_results}.
However, the quality gain after finetuning for YT-ASL based models is often higher than their YT-Full counterparts, where the largest gain reaches $\sim$28 BLEURT for model (5).
We argue that pretraining on YT-ASL mainly teaches understanding of ASL, so pretrained performance on other sign languages is poor, but finetuning can quickly adapt the learned representations to other sign languages.

Note we also finetuned the vanilla ByT5 model without SLT pretraining for reference, which achieves 7.68 and 3.10 BLEU on Elementary23 and WMT23 DSGS SS, respectively.
Despite their inferiority, these results already surpass the previous SOTA, further showing the potential of ByT5.

\textbf{A model's pretraining performance may be misleading when estimating its downstream finetuning performance, depending on the evaluation metric.}
Intuitively, a model with better pretrained results should result in better finetuned results.
The Spearman's correlation results in Table \ref{tab:spearman_corr} confirm this intuition, where the correlation scores are positive across metrics.
However, BLEU and ChrF have a correlation score of 0.347 and 0.186, respectively, which are very moderate.
The correlation for ChrF is even not significant, which may be caused by the use of BLEU as the model selection metric.
In contrast, the correlation of BLEURT reaches 0.578 and is significant at $p<0.01$.

\begin{wraptable}{r}{0.4\textwidth}
\centering
\small
\setlength{\tabcolsep}{4pt}
\small
\begin{tabular}{lccc}
\toprule
& BLEU & ChrF & BLEURT \\
Spearman's $\rho$ & 0.347$^\dag$ & 0.186 & \textbf{0.578}$^\ddag$ \\
\bottomrule
\end{tabular}
\caption{\label{tab:spearman_corr} Spearman correlation between direct (i.e. pretraining) and finetuning SLT results under different metrics based on Tables \ref{tab:downstream_slt_results} and \ref{tab:downstream_slt_pretrain_results}. $^\dag/^\ddag$: significant at $p < 0.05/0.01$.}
\end{wraptable}

\textbf{Model, data and language scaling together leads to new state-of-the-art results.}
Diving deeper into Table \ref{tab:downstream_slt_results}, we see clear improvements brought by scaling model size, data, and/or languages for SLT.
Adding YT-ASL SLT data into the pretraining yields $\sim$10 average BLEURT improvement ((1)$\rightarrow$(2)).
Jointly training SLT with MT data produces another gain of $\sim$6 BLEURT ((2)$\rightarrow$(3)).
Data augmentation adds an improvement of $\sim$3 BLEURT ((3)$\rightarrow$(4)), which matches the quality achieved by adding large amount of extra multilingual SLT data to the baseline, i.e. (4) 40.91 vs. (6) 40.28.
By further increasing the amount of MT and augmented SLT data as well as the ByT5 model size, we reach an average BLEU, ChrF and BLEURT of 16.90, 39.49, and 49.60, respectively (model (8)).
These results also outperform previous best results, establishing the new SOTA.

\textbf{Multilingual finetuning improves multilingual SLT with encouraging performance, although it still underperforms bilingual finetuning on average.}
We next study multilingual finetuning on the direct mix of different SLT benchmarks.
Table \ref{tab:downstream_slt_results} ((8)$\rightarrow$(9)) shows that multilingual SLT outperforms previous SOTA on almost all benchmarks, but underperforms its bilingual counterpart by 1.22 BLEURT on average.
How to balance modeling capacity among different languages in a joint model and avoid cross-lingual/modality interference is a well known issue in multilingual modeling~\citep{arivazhagan2019massively,zhang-etal-2020-improving,Wang_2019_CVPR}, and multilingual SLT also suffers~\citep{9878501}, which we leave to future.
Still, multilingual SLT facilitates transfer to LIS-CH, leading to a substantial gain of 2.6 BLEURT ((8)$\rightarrow$(9)).

\section{Related Work}\label{sec:related_work}

The main bottleneck of SLT is data scarcity.
Early studies address this issue by developing more data efficient neural architectures and/or training algorithms.
\citet{camgoz2018neural} pioneered the study with encoder-decoder based recurrent models for SLT, which was quickly replaced by Transformer and multi-task learning with CTC regularization~\citep{camgoz2020sign}.
\citet{zhou2021spatial} developed spatial-temporal architecture to model the collaboration of different visual cues.
Another way is to transfer the knowledge from pretrained models, augmentations, and other tasks.
\citet{chen2022simple,chen2022two} proposed to leverage pretrained visual encoders and MT models to improve SLT, while \citet{zhang2023sltunet} explored transferring translation knowledge from MT data directly. \citet{zhou2021improving} employed back-translation to generate pseudo SLT training data.
\citet{ye-etal-2023-cross} augmented the training data by the mix-up algorithm.
Yet another way to address data scarcity is to make data less scarce.
\citet{shi-etal-2022-open}, \citet{uthus2024youtube}, and \citet{youtubesl25} collected large-scale SLT data from \youtube and improved data quality via manual filtering; \citet{albanie2021bbc} developed a British Sign Language translation corpus based on BBC broadcasts instead. \citet{fsInSlt} scaled up ASL data by eschewing manual filtering and tolerating misaligned or irrelvant data. We follow and scale to noisy multilingual sign language data, MT data, and augmented paralel data.

Despite the aforementioned advancements, many studies still heavily depend on \textit{sign glosses}.
As a bridge between sign video and target text, sign glosses ease learning, but are expensive to annotate, not always available, nonstandardized, and cannot cope with sign language grammar in generality~\citep{desai2024systemic}.
Recent research therefore turns to gloss-free SLT, which often underperforms gloss-based counterparts~\citep{lin2023gloss,zhou2023gloss,wong2024signgpt} and performs poorly in open-domain settings~\citep{sandoval-castaneda-etal-2023-ttics,xu-etal-2023-knowcomp,muller-etal-2023-findings}.
We substantially improve gloss-free SLT performance across benchmarks through scaling.
In this regard, our work is closely related to SSVP-SLT~\citep{rust2024towards} but with different focuses.
SSVP-SLT improves SLT by pretraining a neural sign encoder through large-scale self-supervised learning.
By contrast, we adopt static landmarks to represent sign frames and improve the translation by transferring knowledge from other languages and tasks.
The methods used in our study are orthogonal to SSVP-SLT.
In addition, our work also falls into the category of improving multilingual SLT~\citep{9878501,gueuwou-etal-2023-jwsign}.
We didn't evaluate our models on these multilingual benchmarks though as they are either unavailable at the time of paper writing or unusable due to licensing issues.

\section{Conclusion, Limitations, and Future Work}\label{sec:conclusion}

We presented a systematic study of data, model and language scaling for SLT via large-scale SLT pretraining.
In general, scaling substantially improves SLT.
We observe positive knowledge transfer from other sign language data and from machine translation data. 
By joint SLT and MT training, we show the feasibility of achieving zero-shot SLT.
Data augmentation expanding SLT data to more spoken languages via off-the-shelf MT models significantly improves multilingual SLT.
Putting everything together, finetuning our pretrained SLT models leads to new state-of-the-art SLT results across 5 benchmarks covering 5 sign languages (but still far from usable quality).

Although our models have nominally been pretrained on a massive number of sign languages (up to 80), we lack comprehensive and reliable multilingual benchmarks to fully understand their abilities and limitations.
In addition, our models are limited to encoder-decoder based (m/By)T5 models, and SLT pretraining requires many computational resources, increasing the difficulty of reproduction.

In the future, we expect that continuing to scale sign language data, number of sign languages, vision pretraining/multimodality, etc. will reap further gains. As suggested by~\citep{fleursasl}, it will be important to evaluate these growing capabilities on multilingual, standardized, open-domain SLT benchmarks.

\section*{Ethics Statement}\label{sec:ethics}

We preprocess all sign videos with simplified landmarks as a form of anonymization and privacy protection.
While the pretraining SLT data is larger scale than prior work, it may still suffer from demographic biases. Even if demographics were represented in proportion to the real world, and even with simplified landmarks, the resulting SLT models may not perform equally across groups and should be evaluated for fairness before real-world deployment.
Our study mainly aims to understand the impact of scaling on SLT, and while we significantly improve translation quality, it is still far from usable for real-world applications. For many such applications, the other half of sign language translation---sign language generation---is also essential, whereas we focus only on sign language understanding in this work. Advancing both of these is critical to ensure that Deaf/Hard of Hearing signers get equal access to technology and the information that comes through it.

\section*{Acknowledgements}

We thank Ankush Garg for valuable feedback on this work, Chris Dyer for constructive comments that greatly improve the quality of this paper, Sam Sepah and Google Translate team for supporting this research.
We also thank the T5X team~\citep{roberts2023scaling} for infrastructure support.

\bibliographystyle{plainnat}
\bibliography{paper}


\appendix

\section{Appendix}

{ 
\setlength{\tabcolsep}{4pt}
\small
\begin{longtable}{clrrrclr}
\caption{\label{tab:lang_stats} Data statistics for \youtube sign language and \madlad spoken language. We list ISO 639-3 code, language name, and the number of hours/clips/videos for sign language; for spoken language, we list BCP-47 code, language name and the number of parallel examples. ``K'': thousand, ``M'': million. Note that like~\citet{youtubesl25} pre-filtering and~\citet{fsInSlt}, these language labels are heuristically estimated based on public video metadata, such as caption language and text in the video title, description, etc.} \\
\toprule
\multicolumn{5}{c}{Sign Language (SL)} & \multicolumn{3}{c}{Spoken Language} \\
\cmidrule(lr){1-5} \cmidrule(lr){6-8}
ISO 639-3 & Name & \# Hours & \# Clips & \# Videos & BCP-47 & Name & \# Examples \\
\cmidrule(lr){1-5} \cmidrule(lr){6-8}
ase & American SL & 2.8K & 285.2K & 25.9K & es & Spanish & 292.8M \\
bzs & Brazilian SL & 590.4 & 60.2K & 5.9K & de & German & 283.3M \\
pso & Polish SL & 421.8 & 41.8K & 3.2K & fr & French & 243.6M \\
ins & Indian SL & 375.3 & 39.3K & 5.7K & it & Italian & 100.1M \\
bfi & British SL & 267.3 & 27.4K & 2.7K & cs & Czech & 53.1M \\
gsg & German SL & 235.2 & 24.2K & 2.1K & pl & Polish & 42.9M \\
fsl & French SL & 193.7 & 19.9K & 2.9K & ru & Russian & 29.0M \\
jsl & Japanese SL & 176.5 & 17.9K & 3.0K & zh & Simplified Chinese & 25.9M \\
ise & Italian SL & 161.0 & 16.8K & 2.4K & ar & Arabic & 18.2M \\
asf & Australian SL & 123.1 & 12.5K & 1.5K & ja & Japanese & 5.3M \\
rsl & Russian SL & 119.4 & 12.1K & 1.2K & hi & Hindi & 1.2M \\
\cmidrule(lr){6-8}
csc & Catalan SL & 114.8 & 11.8K & 1.7K & nl & Dutch & 93.1M \\
csn & Colombian SL & 107.8 & 11.1K & 478.0 & pt & Portuguese & 83.7M \\
aed & Argentine SL & 86.1 & 8.8K & 522.0 & sv & Swedish & 51.9M \\
mfs & Mexican SL & 76.9 & 7.6K & 434.0 & hu & Hungarian & 40.0M \\
kvk & Korean SL & 69.3 & 7.1K & 612.0 & da & Danish & 38.2M \\
hsh & Hungarian SL & 55.9 & 5.8K & 1.3K & fi & Finnish & 34.1M \\
sgg & Swiss German SL & 48.7 & 5.1K & 634.0 & el & Greek & 25.2M \\
prl & Peruvian SL & 42.3 & 4.2K & 147.0 & sk & Slovak & 25.0M \\
fse & Finnish SL & 41.2 & 4.3K & 593.0 & no & Norwegian & 19.4M \\
swl & Swedish SL & 38.7 & 4.0K & 563.0 & bg & Bulgarian & 15.5M \\
asq & Austrian SL & 31.8 & 3.2K & 592.0 & lt & Lithuanian & 15.3M \\
tsm & Turkish SL & 31.8 & 3.3K & 446.0 & lv & Latvian & 14.3M \\
dse & Dutch SL & 31.4 & 3.2K & 352.0 & sl & Slovenian & 11.8M \\
cse & Czech SL & 29.2 & 3.0K & 336.0 & et & Estonian & 11.0M \\
inl & Indonesian SL & 27.5 & 2.8K & 236.0 & ko & Korean & 5.8M \\
nsl & Norwegian SL & 22.0 & 2.2K & 215.0 & hr & Croatian & 5.3M \\
hks & Hong Kong SL & 21.3 & 2.2K & 248.0 & is & Icelandic & 4.1M \\
tss & Taiwan SL & 19.6 & 2.0K & 258.0 & sr & Serbian & 2.5M \\
gss & Greek SL & 18.2 & 1.9K & 169.0 & tr & Turkish & 2.5M \\
dsl & Danish SL & 16.0 & 1.6K & 188.0 & vi & Vietnamese & 1.5M \\
csg & Chilean SL & 15.2 & 1.5K & 184.0 & id & Indonesian & 1.4M \\
sfb & French Belgian SL & 14.4 & 1.5K & 358.0 & he & Hebrew & 1.1M \\
isr & Israeli SL & 14.3 & 1.4K & 289.0 & th & Thai & 1.1M \\
vietnam & Vietnamese SL & 14.2 & 1.1K & 97.0 & ms & Malay & 907.5K \\
isg & Irish SL & 13.2 & 1.4K & 101.0 & uk & Ukrainian & 881.0K \\
slovenia & Slovenian SL & 12.7 & 1.3K & 177.0 & ca & Catalan & 686.2K \\
nzs & New Zealand SL & 12.3 & 1.3K & 224.0 & ta & Tamil & 396.9K \\
icl & Icelandic SL & 11.1 & 1.2K & 213.0 & fil & Filipino & 369.8K \\
sls & Singapore SL & 9.9 & 1.0K & 148.0 & ne & Nepali & 277.9K \\
tsq & Thai SL & 9.0 & 923.0 & 150.0 & cy & Welsh & 93.3K \\
pks & Pakistani SL & 8.6 & 902.0 & 145.0 \\
svk & Slovak SL & 8.5 & 886.0 & 114.0 \\
jos & Jordanian SL & 8.3 & 880.0 & 124.0 \\
lls & Lithuanian SL & 7.6 & 792.0 & 166.0 \\
csr & Costa Rican SL & 7.6 & 789.0 & 45.0 \\
psr & Portuguese SL & 7.4 & 764.0 & 127.0 \\
rms & Romanian SL & 7.3 & 747.0 & 60.0 \\
xml & Malaysian SL & 7.2 & 668.0 & 85.0 \\
ecs & Ecuadorian SL & 7.2 & 727.0 & 28.0 \\
psp & Filipino SL & 6.8 & 715.0 & 85.0 \\
sfs & South African SL & 5.2 & 543.0 & 41.0 \\
ugy & Uruguay SL & 4.7 & 483.0 & 113.0 \\
esn & Salvadoran SL & 3.9 & 403.0 & 15.0 \\
xki & Kenyan SL & 3.8 & 384.0 & 23.0 \\
serbia & Serbian SL & 3.3 & 351.0 & 36.0 \\
csq & Croatian SL & 3.1 & 326.0 & 31.0 \\
esl & Egyptian SL & 3.0 & 262.0 & 32.0 \\
psl & Puerto Rican SL & 1.8 & 181.0 & 17.0 \\
bengladesh & Bengali SL & 1.6 & 165.0 & 15.0 \\
gsm & Guatemalan SL & 1.4 & 144.0 & 26.0 \\
xms & Moroccan SL & 1.2 & 125.0 & 4.0 \\
lsp & Panamanian SL & 1.2 & 123.0 & 8.0 \\
fcs & Quebec SL & 0.9 & 91.0 & 27.0 \\
eso & Estonian SL & 0.9 & 90.0 & 13.0 \\
emirati & UAE SL & 0.8 & 81.0 & 14.0 \\
vsl & Venezuelan SL & 0.7 & 69.0 & 15.0 \\
pys & Paraguayan SL & 0.7 & 69.0 & 5.0 \\
kazakh & Kazakh SL & 0.6 & 69.0 & 6.0 \\
hds & Honduran SL & 0.6 & 67.0 & 8.0 \\
macau & Macau SL & 0.6 & 109.0 & 109.0 \\
sdl & Saudi Arabian SL & 0.5 & 49.0 & 7.0 \\
doq & Dominican SL & 0.5 & 45.0 & 10.0 \\
belarus & Belarusian SL & 0.4 & 29.0 & 5.0 \\
bqn & Bulgarian SL & 0.3 & 31.0 & 9.0 \\
sqs & Sri Lankan SL & 0.3 & 29.0 & 8.0 \\
lsl & Latvian SL & 0.2 & 27.0 & 3.0 \\
bvl & Bolivian SL & 0.2 & 26.0 & 5.0 \\
nsi & Nigerian SL & 0.1 & 12.0 & 2.0 \\
nsp & Nepali SL & 0.1 & 6.0 & 1.0 \\
\bottomrule
\end{longtable}
}

\subsection{Setup}\label{app:model_setup}

\paragraph{Sign Video Preprocessing}

Our landmark preprocessing is identical to~\cite{uthus2024youtube}, and we use the same random 34-second video clipping as~\cite{fleursasl}.
We preprocess sign language video with its default frame rate but discard every second frame for computational efficiency. We convert each video frame to a 255-dimensional normalized vector using MediaPipe Holistic landmarks~\citep{lugaresi2019mediapipe}, which also facilitates video anonymization.
The input video is eventually transformed into a vector sequence and then mapped to the encoder via a linear projection layer.

\paragraph{Downstream Benchmarks}

Note the official SLT track in WMT23 for LIS-CH and LSF-CH is for sign language generation rather than translation.
We reversed it as a SLT dataset.
\fleurs-ASL\#0 is the subset of \fleurs-ASL~\citep{fleursasl} recorded by signer \#0, i.e. 353 sentences from FLORES~\citep{nllb2022} translated into ASL by a Certified Deaf Interpreter. We report only signer \#0 because the rest of the benchmark was not complete when these experiments were run.

For these benchmarks, we only use the sign language video and target text \textit{without} glosses. All SLT models in this study are gloss-free.

\paragraph{Model Setting}

For \textbf{Pretraining}, we use a batch size of 256 and a constant learning rate of 0.001.
We pretrain models up to 1M steps using 64/64/128 TPU-v3 chips for Base/Large/XL, taking 7-20 days. We select the best checkpoint for downstream application based on the How2Sign dev performance measured by BLEU\footnote{We didn't adopt BLEURT for model selection because it's significantly more expensive and time-consuming than BLEU.}.

For \textbf{Finetuning}, we use a batch size of 32 and a constant learning rate of 0.0005.
By default, we perform finetuning on each downstream benchmark separately.
We only consider the SLT task at finetuning, and directly finetune the model on well aligned (sign video segment, target translation) pairs, which is provided in all downstream benchmarks.
We tune models up to 50K steps using 16/32 TPU-v3 chips for Base/XL, taking 2$\sim$5 days. We select the best checkpoint for final evaluation based on the dev-set BLEU.

\subsection{More Results and Analysis}

\begin{figure*}[h!]
\centering
\small

\includegraphics[width=0.85\textwidth]{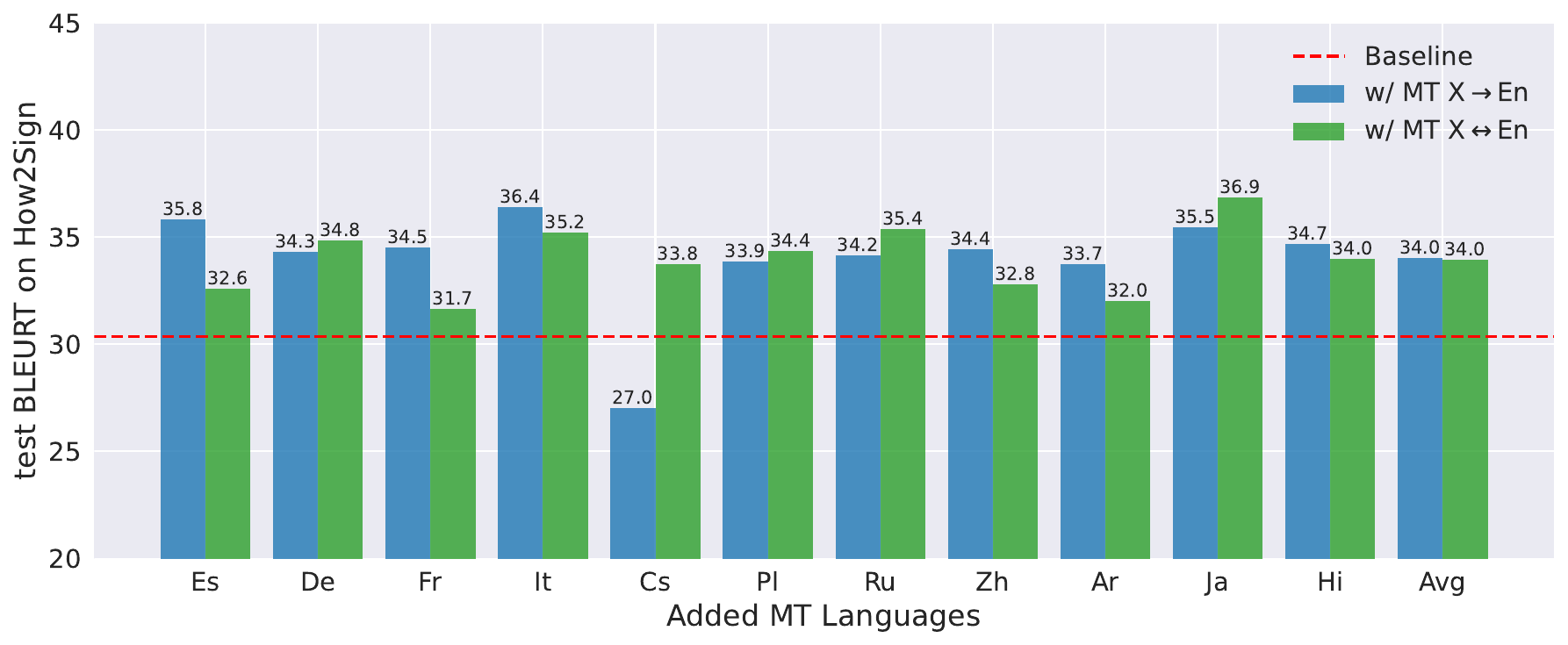}

\caption{\label{fig:mt_transfer_2en_how2sign} Pretraining performance for \textit{Baseline + MT} when varying MT languages on How2Sign test set. We show BLEURT$\uparrow$ results and set $p_{mt}=0.5$. Note only YT-ASL and bilingual MT data are used, i.e. MT languages are added separately instead of jointly. Results are for ByT5 Base. 
``X$\rightarrow$En'': MT data for translation into English; ``X$\leftrightarrow$En'': MT data for both translation directions; ``Avg'': average performance over languages. 
MT languages are arranged in descending order from left to right based on the quantity of translation data available for each language.}
\end{figure*}

\begin{figure*}[h!]
\centering
\small

\includegraphics[width=0.8\textwidth]{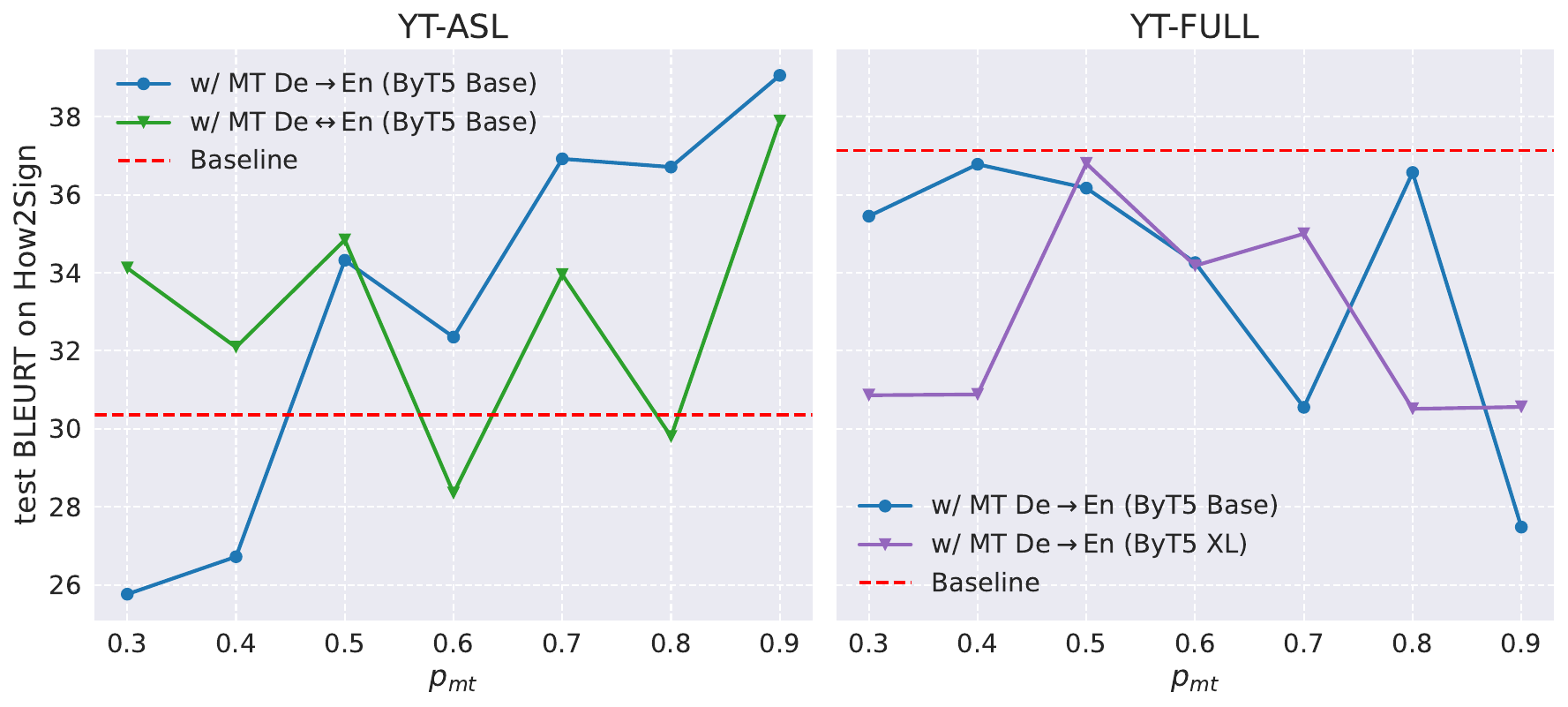}

\caption{\label{fig:mt_transfer_ratio_how2sign} Pretraining performance for \textit{Baseline + MT} when changing the mixing ratio of MT data $p_{mt}$ on How2Sign test set. We show BLEURT$\uparrow$ results and vary $p_{mt}$ from 0.3 to 0.9. Note only bilingual MT En-De data are explored.}
\end{figure*}

\begin{figure*}[h!]
\centering
\small

\subcaptionbox{\label{fig:mt_transfer_m4_small} Results for training with MT-Small.}{
\includegraphics[width=0.8\textwidth]{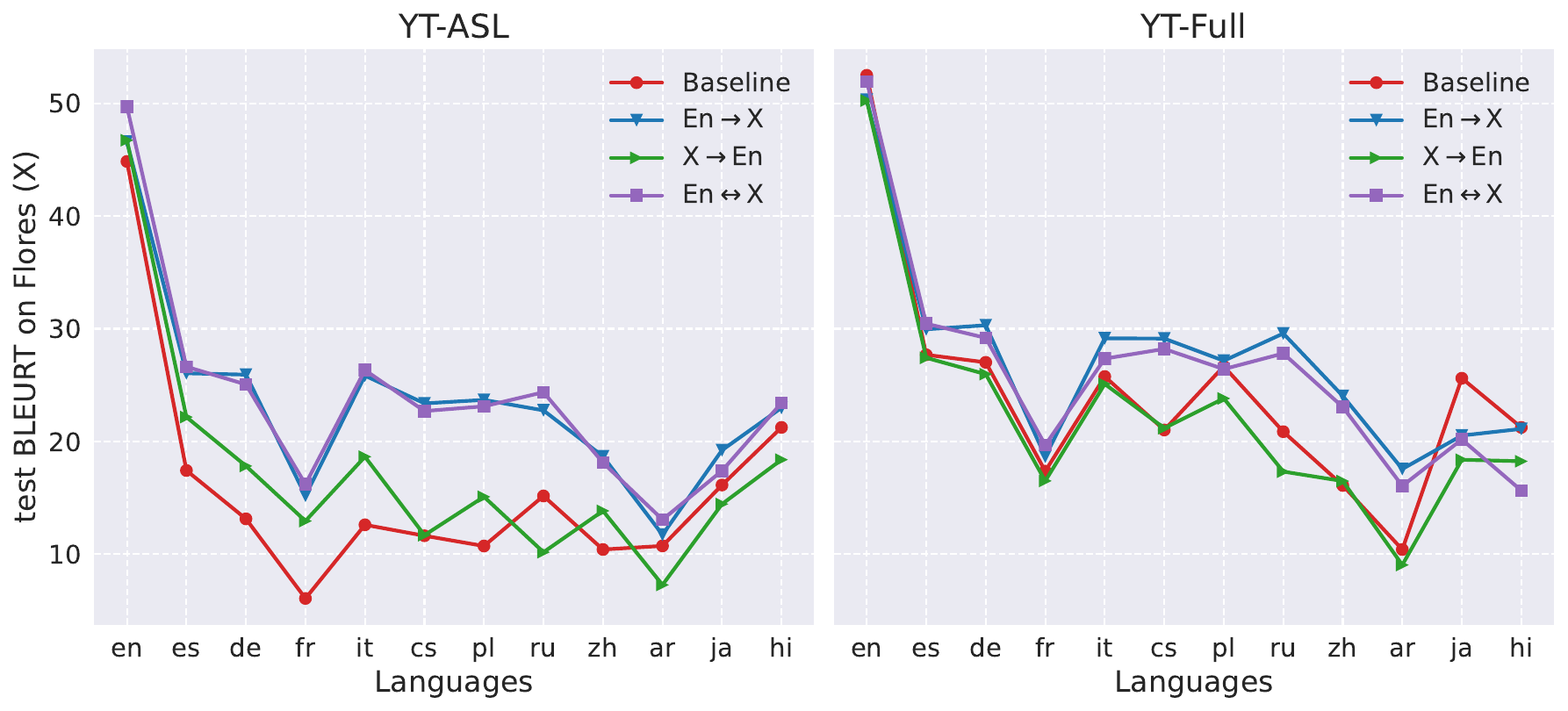}
}

\vspace{\baselineskip}

\subcaptionbox{\label{fig:mt_transfer_m4_large} Results for training with MT-Large.}{
\includegraphics[width=0.8\textwidth]{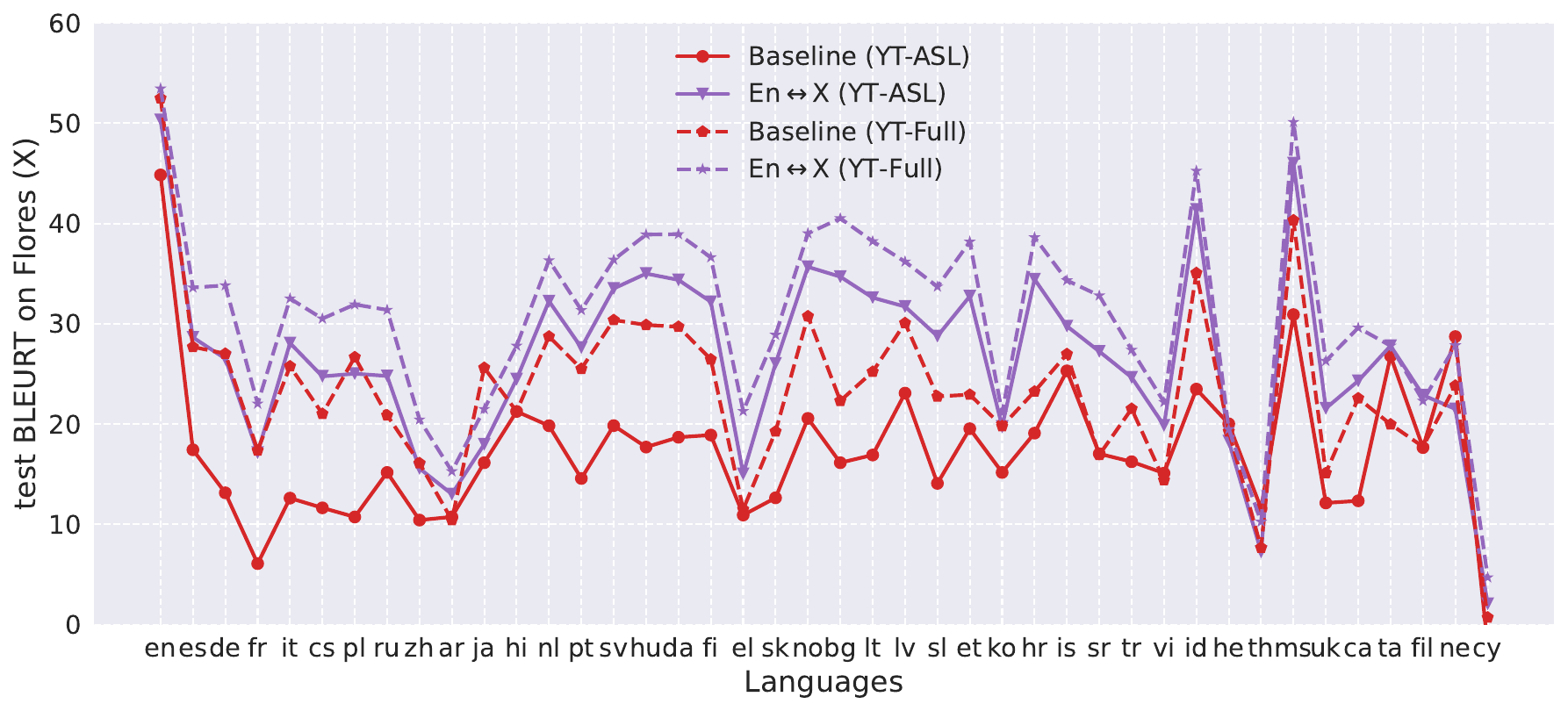}
}

\caption{\label{fig:mt_transfer_m4lang} Per-language pretraining performance for \textit{Baseline + MT} with $p_{mt}=0.9$ when scaling up languages and data. We show BLEURT$\uparrow$ results on \fleurs-ASL\#0. We add multilingual MT data into SLT pretraining and compare MT-Small with MT-Large. Results are for ByT5 Base.}
\end{figure*}

\begin{figure*}[h!]
\centering
\small

\subcaptionbox{\label{fig:mt_transfer_m4_small_distil} Results for training with MT-Small and Aug-YT-ASL/Full-Small.}{
\includegraphics[width=0.8\textwidth]{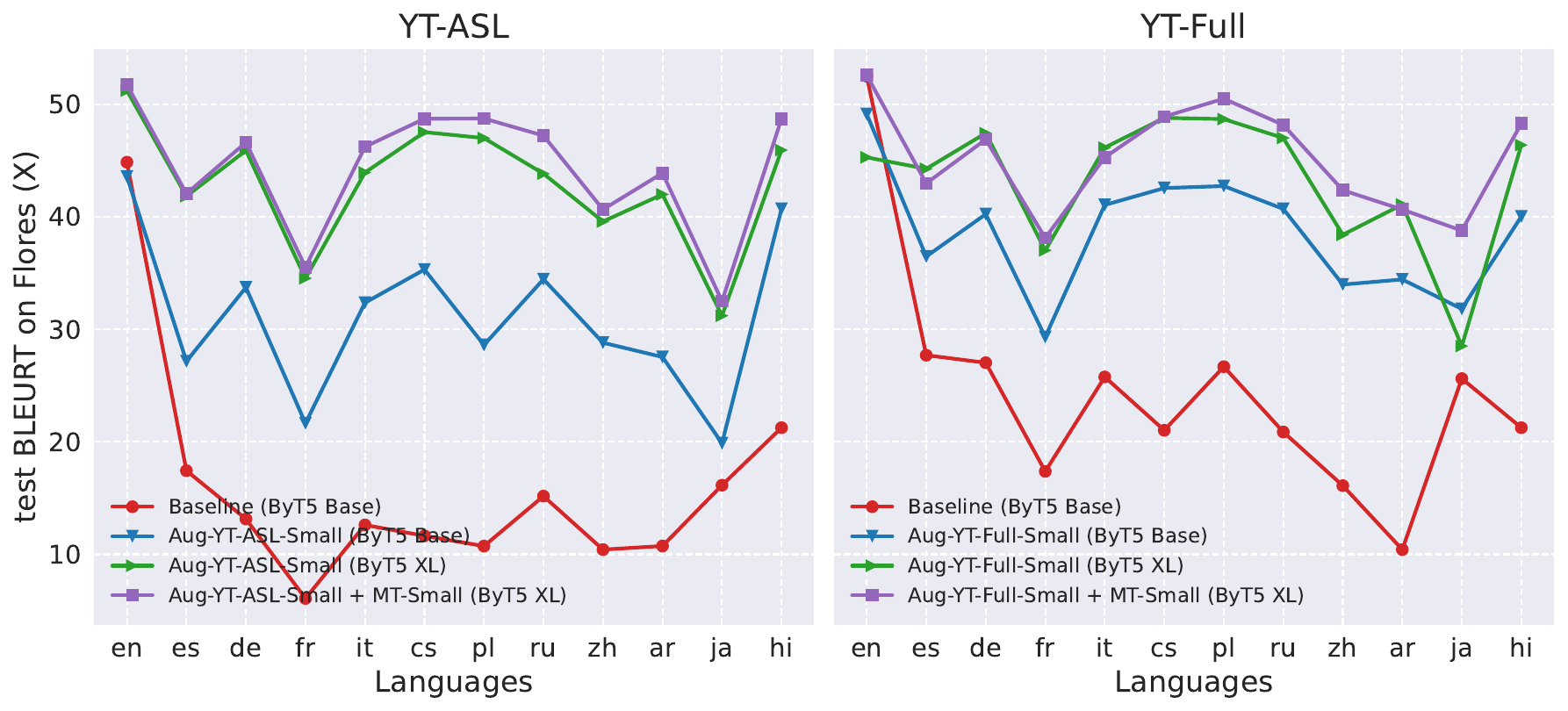}
}

\vspace{\baselineskip}

\subcaptionbox{\label{fig:mt_transfer_m4_large_distil} Results for training with MT-Large and Aug-YT-ASL/Full-Large.}{
\includegraphics[width=0.8\textwidth]{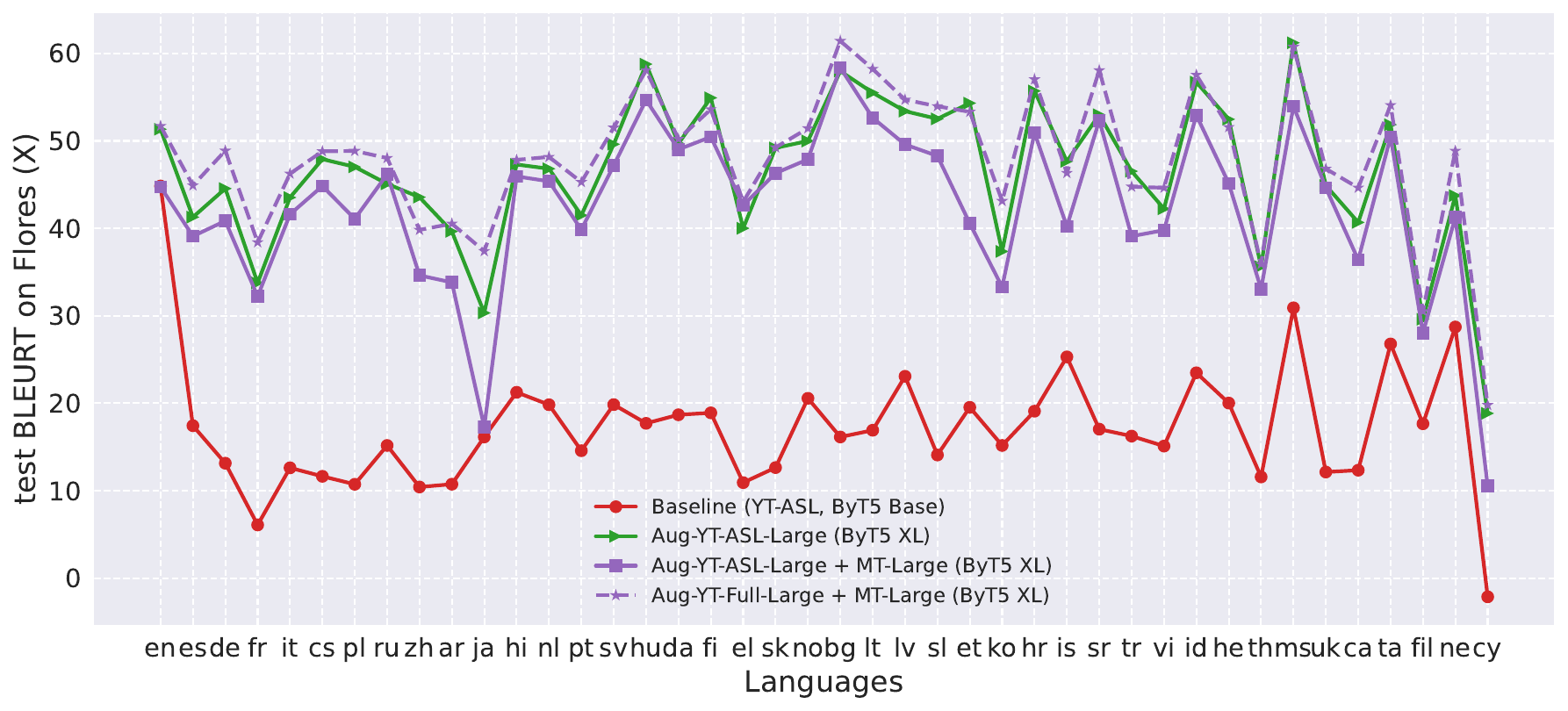}
}

\caption{\label{fig:mt_transfer_distil} Per-language pretraining performance for SLT with augmented SLT data. We show BLEURT$\uparrow$ results for \textit{Baseline + Augmented SLT + MT} with $p_{mt}=0.9$ on \fleurs test set. MT data are multilingual in both directions. Data augmentation substantially improves SLT performance across languages.}
\end{figure*}

\begin{table*}[h!]
\centering
\setlength{\tabcolsep}{4pt}
\small

\subcaptionbox{\label{tab:slt_bleu} BLEU$\uparrow$ scores.}{
\resizebox{0.85\textwidth}{!}{\begin{tabular}{llrrrrrrr}
\toprule
\multirow{2}{*}{ID} & \multirow{2}{*}{Model} & \multirow{2}{*}{H2S} & \multirow{2}{*}{E23} & \multicolumn{4}{c}{WMT23} & \multirow{2}{*}{Avg} \\
\cmidrule(lr){5-8}
& & & & LIS-CH & LSF-CH & SRF & SS & \\
\cmidrule(lr){1-9}
0 & Prevous SOTA & 18.10 & 5.69 & 5.20 & 7.00 & 0.30 & 7.50 & 7.30 \\
\cmidrule(lr){1-9}
1 & ByT5 Base & 3.71 & 7.68 & 0.40 & 0.79 & 1.01 & 3.10 & 2.78 \\
2 & 1 + Baseline + YT-ASL & 17.94 & 16.58 & 6.59 & 8.65 & 2.14 & 8.69 & 10.10 \\
3 & 2 + MT-Small ($p_{mt}=0.9$) & 17.30 & 17.90 & 8.84 & 11.86 & 2.04 & 11.01 & 11.49 \\
4 & 3 + Aug-YT-ASL-Small & 18.55 & 22.09 & 9.81 & 12.91 & 2.90 & 14.34 & 13.43 \\
5 & 4 + Aug-YT-ASL\&MT-Large + ByT5 XL & 19.31 & 24.08 & 9.70 & 11.71 & 2.76 & 13.26 & 13.47 \\
\cmidrule(lr){1-9}
6 & 2 + YT-Full & 19.79 & 21.81 & 12.99 & 15.32 & 2.07 & 13.71 & 14.28 \\
7 & 6 + Aug-YT-ASL\&MT-Small & 18.98 & 23.60 & 13.63 & 15.75 & 2.85 & 14.44 & 14.88 \\
8 & 7 + Aug-YT-ASL\&MT-Large + ByT5 XL & \textbf{21.06} & \textbf{25.65} & 14.93 & \textbf{18.77} & 2.80 & \textbf{18.17} & \textbf{16.90} \\
\cmidrule(lr){1-9}
9 & 8 + Multilingual SLT Tuning & 19.25 & 23.05 & \textbf{16.79} & 17.22 & \textbf{2.91} & 15.92 & 15.86 \\
\bottomrule
\end{tabular}}
}
\vspace{\baselineskip}

\subcaptionbox{\label{tab:slt_chrf} ChrF$\uparrow$ scores.}{
\resizebox{0.85\textwidth}{!}{\begin{tabular}{llrrrrrrr}
\toprule
\multirow{2}{*}{ID} & \multirow{2}{*}{Model} & \multirow{2}{*}{H2S} & \multirow{2}{*}{E23} & \multicolumn{4}{c}{WMT23} & \multirow{2}{*}{Avg} \\
\cmidrule(lr){5-8}
& & & & LIS-CH & LSF-CH & SRF & SS & \\
\cmidrule(lr){1-9}
0 & Prevous SOTA & - & - & - & - & 17.50 & - & - \\
\cmidrule(lr){1-9}
1 & ByT5 Base & 19.55 & 25.19 & 14.81 & 15.56 & 11.08 & 21.75 & 17.99 \\
2 & 1 + Baseline + YT-ASL & 38.78 & 41.34 & 27.57 & 29.99 & 17.06 & 38.31 & 31.18 \\
3 & 2 + MT-Small ($p_{mt}=0.9$) & 39.01 & 45.40 & 30.56 & 33.67 & 16.13 & 40.38 & 34.19 \\
4 & 3 + Aug-YT-ASL-Small & 39.64 & 47.92 & 31.40 & 34.95 & 17.49 & 43.56 & 35.83 \\
5 & 4 + Aug-YT-ASL\&MT-Large + ByT5 XL & 40.15 & 48.70 & 29.44 & 33.20 & 17.96 & 40.21 & 34.94 \\
\cmidrule(lr){1-9}
6 & 2 + YT-Full & 41.13 & 47.71 & 38.23 & 38.13 & 16.78 & 42.33 & 37.38 \\
7 & 6 + Aug-YT-ASL\&MT-Small & 40.35 & 49.57 & 38.07 & 39.88 & 19.34 & 42.83 & 38.34 \\
8 & 7 + Aug-YT-ASL\&MT-Large + ByT5 XL & \textbf{41.97} & \textbf{50.09} & 38.90 & \textbf{40.43} & \textbf{19.58} & \textbf{45.94} & \textbf{39.49} \\
\cmidrule(lr){1-9}
9 & 8 + Multilingual SLT Tuning & 40.39 & 48.49 & \textbf{40.28} & 38.63 & 18.72 & 45.22 & 38.62 \\
\bottomrule
\end{tabular}}
}
\vspace{\baselineskip}

\subcaptionbox{\label{tab:bleurt} BLEURT$\uparrow$ scores.}{
\resizebox{0.85\textwidth}{!}{\begin{tabular}{llrrrrrrr}
\toprule
\multirow{2}{*}{ID} & \multirow{2}{*}{Model} & \multirow{2}{*}{H2S} & \multirow{2}{*}{E23} & \multicolumn{4}{c}{WMT23} & \multirow{2}{*}{Avg} \\
\cmidrule(lr){5-8}
& & & & LIS-CH & LSF-CH & SRF & SS & \\
\cmidrule(lr){1-9}
0 & Prevous SOTA & 50.80 & - & 25.20 & 18.80 & 24.60 & 37.70 & - \\
\cmidrule(lr){1-9}
1 & ByT5 Base & 34.00 & 22.14 & 22.77 & 7.74 & 15.41 & 26.88 & 21.49 \\
2 & 1 + Baseline + YT-ASL & 51.74 & 37.79 & 24.24 & 15.43 & 21.82 & 35.59 & 31.10 \\
3 & 2 + MT-Small ($p_{mt}=0.9$) & 52.62 & 45.98 & 33.10 & 24.58 & 23.33 & 45.45 & 37.51 \\
4 & 3 + Aug-YT-ASL-Small & 53.36 & 49.34 & 38.61 & 28.70 & 25.87 & 49.61 & 40.91 \\
5 & 4 + Aug-YT-ASL\&MT-Large + ByT5 XL & 54.28 & 54.16 & 38.93 & 27.29 & 28.42 & 51.73 & 42.47 \\
\cmidrule(lr){1-9}
6 & 2 + YT-Full & 53.51 & 49.48 & 42.11 & 31.16 & 21.15 & 44.28 & 40.28 \\
7 & 6 + Aug-YT-ASL\&MT-Small & 53.70 & 53.13 & 45.09 & 37.69 & 30.31 & 52.45 & 45.40 \\
8 & 7 + Aug-YT-ASL\&MT-Large + ByT5 XL & \textbf{55.69} & \textbf{56.94} & 51.94 & \textbf{41.14} & \textbf{33.94} & 57.96 & \textbf{49.60} \\
\cmidrule(lr){1-9}
9 & 8 + Multilingual SLT Tuning & 53.47 & 55.57 & \textbf{54.54} & 39.26 & 29.33 & \textbf{58.08} & 48.38 \\
\bottomrule
\end{tabular}}
}
\caption{\label{tab:downstream_slt_results_full} Finetuning performance on downstream SLT benchmarks. ``H2S/E23'': How2Sign/Elementary23. ``SRF/SS'': WMT23 DSGS SRF/SS test split. ``Avg'': averaged performance over all benchmarks. MT data are added in both translation directions. Previous SOTA: How2Sign~\citep{fsInSlt}, Elementary23~\citep{voskou2023new} and WMT23 SRF~\citep{muller-etal-2023-findings}, WMT23 LIS-CH, LSF-CH, SS~\citep{youtubesl25}. Scaling SLT reaches new SOTA across benchmarks. \textit{All models are finetuned on each SLT benchmark separately except (9).}}
\end{table*}

\begin{table*}[h!]
\centering
\setlength{\tabcolsep}{4pt}
\small

\subcaptionbox{\label{tab:p_slt_bleu} BLEU$\uparrow$ scores.}{
\resizebox{0.85\textwidth}{!}{\begin{tabular}{llrrrrrrr}
\toprule
\multirow{2}{*}{ID} & \multirow{2}{*}{Model} & \multirow{2}{*}{H2S} & \multirow{2}{*}{E23} & \multicolumn{4}{c}{WMT23} & \multirow{2}{*}{Avg} \\
\cmidrule(lr){5-8}
& & & & LIS-CH & LSF-CH & SRF & SS & \\
\cmidrule(lr){1-9}
1 & ByT5 Base &  \\
2 & 1 + Baseline + YT-ASL & 3.77 & 0.06 & 0.15 & 0.35 & 0.15 & 0.15 & 0.77 \\
3 & 2 + MT-Small ($p_{mt}=0.9$) & 4.75 & 0.02 & 0.07 & 0.25 & 0.06 & 0.12 & 0.88 \\
4 & 3 + Aug-YT-ASL-Small & 3.31 & 0.01 & 0.12 & 0.43 & 0.12 & 0.31 & 0.72 \\
5 & 4 + Aug-YT-ASL\&MT-Large + ByT5 XL & 2.81 & 0.21 & 0.22 & 0.31 & 0.06 & 0.24 & 0.64 \\
\cmidrule(lr){1-9}
6 & 2 + YT-Full & 5.78 & 0.33 & 3.43 & 5.69 & 1.08 & 3.88 & 3.37 \\
7 & 6 + Aug-YT-ASL\&MT-Small & 4.10 & 0.05 & 1.67 & 2.65 & 0.50 & 2.21 & 1.86 \\
8 & 7 + Aug-YT-ASL\&MT-Large + ByT5 XL & 4.05 & 2.45 & 4.50 & 3.73 & 0.64 & 3.45 & 3.14 \\
\bottomrule
\end{tabular}}
}
\vspace{\baselineskip}

\subcaptionbox{\label{tab:p_slt_chrf} ChrF$\uparrow$ scores.}{
\resizebox{0.85\textwidth}{!}{\begin{tabular}{llrrrrrrr}
\toprule
\multirow{2}{*}{ID} & \multirow{2}{*}{Model} & \multirow{2}{*}{H2S} & \multirow{2}{*}{E23} & \multicolumn{4}{c}{WMT23} & \multirow{2}{*}{Avg}  \\
\cmidrule(lr){5-8}
& & & & LIS-CH & LSF-CH & SRF & SS & \\
\cmidrule(lr){1-9}
1 & ByT5 Base &  \\
2 & 1 + Baseline + YT-ASL & 20.65 & 6.62 & 12.67 & 15.1 & 13.89 & 13.95 & 13.81 \\
3 & 2 + MT-Small ($p_{mt}=0.9$) & 19.55 & 0.05 & 10.9 & 11.33 & 10.75 & 12.89 & 10.91 \\
4 & 3 + Aug-YT-ASL-Small & 15.21 & 0.05 & 9.67 & 9.87 & 5.89 & 14.28 & 9.16 \\
5 & 4 + Aug-YT-ASL\&MT-Large + ByT5 XL & 11.40 & 9.25 & 8.75 & 6.88 & 3.35 & 7.55 & 7.86 \\
\cmidrule(lr){1-9}
6 & 2 + YT-Full & 23.44 & 14.81 & 25.34 & 26.78 & 16.42 & 28.36 & 22.53 \\
7 & 6 + Aug-YT-ASL\&MT-Small & 18.18 & 10.22 & 19.81 & 22.64 & 10.25 & 25.95 & 17.84 \\
8 & 7 + Aug-YT-ASL\&MT-Large + ByT5 XL & 13.47 & 22.34 & 25.53 & 26.16 & 14.82 & 26.96 & 21.55 \\
\bottomrule
\end{tabular}}
}
\vspace{\baselineskip}

\subcaptionbox{\label{tab:p_slt_bleurt} BLEURT$\uparrow$ scores.}{
\resizebox{0.85\textwidth}{!}{\begin{tabular}{llrrrrrrr}
\toprule
\multirow{2}{*}{ID} & \multirow{2}{*}{Model} & \multirow{2}{*}{H2S} & \multirow{2}{*}{E23} & \multicolumn{4}{c}{WMT23} & \multirow{2}{*}{Avg}  \\
\cmidrule(lr){5-8}
& & & & LIS-CH & LSF-CH & SRF & SS & \\
\cmidrule(lr){1-9}
1 & ByT5 Base &  \\
2 & 1 + Baseline + YT-ASL & 30.36 & 9.13 & 9.32 & 6.33 & 9.69 & 10.45 & 12.55 \\
3 & 2 + MT-Small ($p_{mt}=0.9$) & 34.24 & 1.07 & 16.38 & 10.44 & 13.14 & 14.81 & 15.01 \\
4 & 3 + Aug-YT-ASL-Small & 25.2 & 1.63 & 23.41 & 10.82 & 10.65 & 22.06 & 15.61 \\
5 & 4 + Aug-YT-ASL\&MT-Large + ByT5 XL & 23.87 & 10.35 & 21.58 & 6.75 & 7.04 & 15.96 & 14.26 \\
\cmidrule(lr){1-9}
6 & 2 + YT-Full & 37.13 & 15.08 & 28.92 & 18.33 & 17.87 & 29.66 & 24.50 \\
7 & 6 + Aug-YT-ASL\&MT-Small & 25.25 & 12.68 & 33.54 & 24.48 & 10.66 & 33.91 & 23.42 \\
8 & 7 + Aug-YT-ASL\&MT-Large + ByT5 XL & 22.41 & 34.14 & 43.07 & 34.26 & 21.52 & 39.47 & 32.48 \\
\bottomrule
\end{tabular}}
}
\caption{\label{tab:downstream_slt_pretrain_results} Pretraining performance on downstream SLT benchmarks.
}
\end{table*}

\paragraph{Different evaluation metrics may disagree.}

There is a hot debate in MT community regarding which metric we should use for translation evaluation~\citep{kocmi-etal-2021-ship}.
While BLEU has been widely adopted, it often shows poor correlation with human evaluation, particularly when the translation models are strong~\citep{ma-EtAl:2019:WMT}.
Instead, neural metrics are recommended~\citep{freitag-etal-2022-results}.
We follow this trend and adopt BLEURT as the main metric.
To be compatible with past studies and also ease future comparison, we also add BLEU and ChrF.
Table \ref{tab:downstream_slt_results_full} shows some disagreements between BLEURT and BLEU/ChrF. 
For example, model (4) performs better than (comparable to) model (5) on average based on ChrF (BLEU), while BLEURT scores show a clear superiority of model (5) over (4).
Evaluation metric selection should be more careful due to these disagreements.
In this study, we rely more on BLEURT for the analysis as it correlates better with human evaluation~\citep{freitag-etal-2022-results}.



\end{document}